\documentclass[sigconf,balance=false]{acmart}
\usepackage{popets}
\setcopyright{popets}
\copyrightyear{YYYY}
\acmYear{YYYY}
\acmVolume{YYYY}
\acmNumber{X}
\acmDOI{XXXXXXX.XXXXXXX}
\acmISBN{}
\acmConference{Proceedings on Privacy Enhancing Technologies}
\usepackage[english]{babel}
\usepackage{bm}

\usepackage{amssymb}
\usepackage{subcaption}
\usepackage{algorithm}
\usepackage{algpseudocode} % for pseudocode
\usepackage{booktabs}
\usepackage{multirow}
\usepackage{subcaption}
\usepackage{amsmath}
\usepackage{graphicx}
\usepackage{listings}
\usepackage{placeins} % in preamble
\title{When Tables Leak: Attacking String Memorization in LLM-Based Tabular Data Generation}
\author{Joshua Ward}
\affiliation{
  \institution{University of California Los Angeles}
  \city{} \state{} \country{} 
}

\author{Bochau Gu}
\affiliation{
  \institution{University of California Los Angeles}
  \city{} \state{} \country{} 
}

\author{Chi-Hua Wang}
\affiliation{
  \institution{University of California Los Angeles}
  \city{} \state{} \country{} 
}

\author{Guang Cheng}
\affiliation{
  \institution{University of California Los Angeles}
  \city{} \state{} \country{} 
}

\begin{document}

\begin{abstract}
Large Language Models (LLMs) have recently demonstrated remarkable performance in generating high-quality tabular synthetic data. In practice, two primary approaches have emerged for adapting LLMs to tabular data generation: (i) fine-tuning smaller models directly on tabular datasets, and (ii) prompting larger models with examples provided in context. In this work, we show that popular implementations from both regimes exhibit a tendency to compromise privacy by reproducing memorized patterns of numeric digits from their training data. To systematically analyze this risk, we introduce a simple No-box Membership Inference Attack (MIA) called LevAtt that assumes adversarial access to only the generated synthetic data and targets the string sequences of numeric digits in synthetic observations. Using this approach, our attack exposes substantial privacy leakage across a wide range of models and datasets, and in some cases, is even a perfect membership classifier on state-of-the-art models. Our findings highlight a unique privacy vulnerability of LLM-based synthetic data generation and the need for effective defenses. To this end, we propose two methods, including a novel sampling strategy that strategically perturbs digits during generation. Our evaluation demonstrates that this approach can defeat these attacks with minimal loss of fidelity and utility of the synthetic data. \footnote{A code repository is available \href{https://github.com/BillGuBochao/LLM_LEVATTACK}{here.}}
\end{abstract}

\maketitle

\section{Introduction}
Machine learning systems across diverse domains—from healthcare databases to financial risk assessment platforms—rely heavily on structured tabular datasets \citep{Giuffr2023HarnessingTP, Assefa, Flanagan22}. This widespread dependence has driven significant interest in synthetic tabular data generation, where computational models learn to produce artificial records that statistically mirror the patterns of real datasets while avoiding direct replication \citep{Fonseca23}. The utility of synthetic data stems from two primary advantages: it can supplement limited training samples to improve model performance on underrepresented populations or rare events, and it facilitates private data sharing by generating records that do not directly correspond to actual individuals. These benefits are especially valuable in privacy-sensitive sectors such as medicine \citep{VALLEVIK2024105413} and banking \citep{Wu2023}, where regulatory constraints often limit access to original data. As a result, synthetic data generation has emerged as an essential technique for expanding machine learning capabilities in data-constrained and privacy-regulated environments\citep{platzer2021holdout,McKenna24}.

Large Language Models (LLMs) have recently emerged as state-of-the-art tabular synthetic data generators. Unlike traditional approaches—such as Generative Adversarial Networks (GANs), Variational Autoencoders (VAEs), and diffusion models—that operate directly over original feature space, LLMs encode tabular rows into string representations and leverage two primary methodologies for generation. In-Context Learning (ICL) approaches include existing tabular rows within the LLM's context window and prompt the LLM to generate additional rows following the observed patterns \citep{cllm2024, kim2024epic, Hollmann2025, ma2024tabpfgentabulardata}, while Supervised Fine-Tuning (SFT) methods train LLMs on larger quantities of string-encoded tabular data to learn the underlying distribution before sampling synthetic records \citep{borisov2023languagemodelsrealistictabular,realtabformer, Harmonic}. Both approaches generate synthetic data by producing new string sequences that are subsequently decoded back into tabular format. In both cases, LLM-based generators have been shown to produce synthetic data that exhibits superior statistical fidelity to original datasets and maintains high utility for downstream machine learning tasks. Moreover, initial evaluations suggest these methods may offer enhanced privacy protection relative to conventional approaches \citep{realtabformer}.

However, these architectural differences raise unresolved questions about privacy. Extensive research has demonstrated that LLMs exhibit tendencies to memorize training data, particularly when exposed to repeated patterns \citep{carlini2021extractingtrainingdatalarge, Kandpal2022DeduplicatingTD}, longer sequences \citep{carlini2023quantifyingmemorizationneurallanguage,wang-etal-2024-unlocking}, or through supervised fine-tuning processes \citep{chu2025sft}. These memorization behaviors, which can be beneficial for language modeling tasks, present unique privacy risks in the context of tabular data generation where training data often contain structured, long sequences of repeated values across many observations.

In order to measure privacy risk, Membership Inference Attacks (MIAs) \citep{Shokri,Carlini2021MembershipIA} have emerged as the linchpin of privacy auditing for tabular synthetic data generators, serving as the primary tool for evaluating privacy risks across diverse generative approaches \citep{ganleaks,Hilprecht2019MonteCA, vanbreugel2023membership,ward2024dataplagiarismindexcharacterizing,Gen-LRA}. However, these methods focus exclusively on the feature space over which traditional generative models operate, potentially missing the string-space vulnerabilities introduced by LLM-based generation entirely.

To bridge this gap, we examine privacy risks in LLM-based tabular data generation by introducing LevAtt, an MIA that exploits Levenshtein Distance on the string representations of tabular data—the actual format LLMs generate—rather than the feature space alone. We find through extensive experimentation in both the ICL and SFT regimes that state-of-the-art LLMs often memorize and replicate numeric values from training data, exposing sensitive information digit-for-digit in synthetic outputs. Even under a conservative no-box threat model, where we assume only adversarial access to the synthetic data, we uncover that LLMs can leak private data through memorized digit patterns, revealing vulnerabilities that conventional feature-space MIAs fail to detect.

To address this new-found risk, we study two defenses: an ad-hoc post-processing algorithm we call Digit Modifier (DM) that flips digits to break sequential patterns and a novel Tendency-based Logit Processor (TLP) that strategically perturbs digits at sample time. We find that both strategies can defeat these attacks, however TLP can effectively control for privacy leakage with minimal effect on the statistical fidelity or downstream machine learning utility of the synthetic data.

\section{Related Work}
To situate LevAtt within the broader landscape of synthetic data research and privacy auditing, we review prior work on tabular data generation, LLM-based modeling, and membership inference attacks, highlighting how LLMs introduce fundamentally new vulnerabilities not present in conventional generators.
\subsection{Tabular Data Generation}
Tabular generative models aim to learn a generator $G$ from training data $T$ that approximates the true data distribution $p_X(X)$. We represent tabular data as a matrix $\mathbf{X} \in \mathcal{X}^{n \times d}$, where $n$ denotes the number of samples, $d$ the number of features, and $\mathcal{X}$ is the feature value domain. Each row $\mathbf{x}_i \in \mathcal{X}^d$ represents a data point drawn from the underlying distribution $p_X(X)$, while columns correspond to features that may have heterogeneous data types. We denote the $j$-th feature value of the $i$-th sample as $\mathbf{x}_{i,j}$. The training dataset $T = \{\mathbf{x}_1, \mathbf{x}_2, \ldots, \mathbf{x}_n\}$ comprises $n$ independent samples from $p_X(X)$. The learned model generates synthetic samples $\tilde{\mathbf{x}} \sim G$, forming a synthetic dataset $S = \{\tilde{\mathbf{x}}_1, \tilde{\mathbf{x}}_2, \ldots, \tilde{\mathbf{x}}_m\}$. In this work, we assume that features can be of a continuous, ordinal, or categorical type.

\textbf{Deep Learning-Based Generation.} In recent years, a variety of conventional tabular synthetic data generators have been proposed including Generative Adversarial Networks \citep{yoon2018pategan, yoon2020anonymization,Xu2019ModelingTD}, likelihood-based methods \cite{Ankan2015, durkan2019neural, pmlr-v206-watson23a}, and diffusion models \cite{tabddpm, autodiff, tabsyn} Each of these operate directly over the feature space $\mathcal{X}^d$, learning mappings $\mathcal{G}: \mathcal{Z} \rightarrow \mathcal{X}^d$ that model the joint distribution $p_X(X)$. Crucially, these approaches fundamentally treat each sample as an atomic unit, generating a complete feature vector $\mathbf{x} \in \mathcal{X}^d$ simultaneously.

\textbf{LLM-Based Generation} In contrast, LLM-based approaches reframe tabular generation as autoregressive text generation. Training samples are encoded into strings, tokenized into sequences from vocabulary $\mathcal{W}$, and the LLM generates according to $p(t) = \prod_{k=1}^{j} p(w_k|w_1, \ldots, w_{k-1})$. Rather than modeling the joint distribution $p_X(X)$ directly, this approach decomposes it through sequential conditioning, generating feature values $x_{i,j}$ token by token based on previously generated values. This sequential generation process fundamentally breaks the atomic unit assumption of other architectures, introducing new dynamics where generated tokens influence later ones through the autoregressive chain.

Within this paradigm, two complementary generation strategies have emerged based on data availability and computational resources. In-context-based methods leverage large foundation models by presenting tabular examples directly within the context window, enabling few-shot generation without parameter updates in low-data regimes \citep{cllm2024, kim2024epic, Hollmann2025, ma2024tabpfgentabulardata}. When larger datasets are available, SFT-based methods instead fine-tune smaller language models through direct optimization on tabular generation tasks \citep{borisov2023languagemodelsrealistictabular, realtabformer, Harmonic}. Both strategies maintain the core autoregressive formulation while differing in how they leverage available data and computational resources.

\subsection{Membership Inference Attacks on Synthetic Tabular Data}
Membership Inference Attacks (MIAs) aim to classify whether a specific observation was a member of the original dataset used to train a model \citep{Shokri,ganleaks,Carlini2021MembershipIA}. Given the generative model $G$ trained on dataset $T$ as defined above, which generates synthetic dataset $S$, an adversary $\mathcal{A}: X \to \{0, 1\}$ aims to determine if a test sample $x^*$ is an element of $T$. Formally, this classification or MIA can be expressed as:
\begin{equation}\label{eq:membership_prediction}
    \mathcal{A}(x^{\star}) = \mathbb{I}\left[f(x^{\star}) > \gamma\right]
\end{equation}
where $\mathbb{I}$ is the indicator function, $f(x^{\star})$ is a scoring function of the test observation $x^*$, and $\gamma$ is an adjustable decision threshold. The success of the attack can be measured using traditional binary classification metrics and can be interpreted as a measure of privacy leakage from a model of the training data.

MIAs have become a mainstay tool to audit the privacy of tabular synthetic data as they represent a material privacy risk associated with the output of a model \citep{houssiau2022tapas,Synth-MIA}. Indeed, a number of distance \citep{ganleaks, ward2024dataplagiarismindexcharacterizing}, density \citep{Hilprecht2019MonteCA, vanbreugel2023membership}, and likelihood-based \citep{groundhog, Meeus_2024,Gen-LRA} attacks have been proposed under a wide variety of threat models. However, these existing methods all attack over the tabular feature space $\mathcal{X}^{d}$. LLM-based generators introduce a fundamentally different vulnerability: they generate in an intermediate string representation space before parsing to a tabular format. This autoregressive string generation process creates a novel attack surface that bypasses traditional feature-space-targeting attacks.
\begin{figure*}
    \centering
    \includegraphics[width=0.75\linewidth]{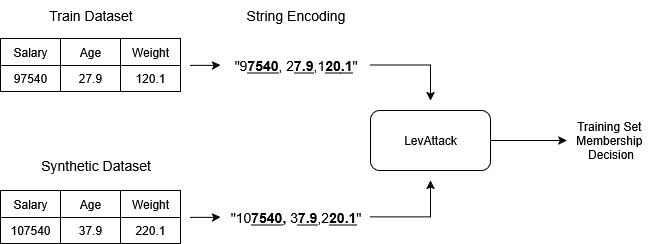}
    \caption{Diagram of Levenshtein Attack. We simply encode rows of tabular data into a string representation from which to attack. LevAtt finds signal in the highly constrained and often duplicated sequences of digits in synthetic tabular data generated by LLMs. In bold and underline: copied sequences of such patterns. Where these rows would be relatively far in Euclidean distance, repeated sequences in their string representations are the source of LevAtt's adversarial advantage.}
    \Description{A diagram illustrating the Levenshtein Attack methodology. The figure shows how tabular data rows are encoded into string representations, with bold and underlined text highlighting copied sequences of digit patterns that reveal synthetic data generated by LLMs. The diagram demonstrates how these repeated sequences provide adversarial advantage despite large Euclidean distances between rows.}
    \label{fig:levattdiagram}
\end{figure*}
\section{Attacking Implied Strings of LLM-Based Generators}

LLM-based tabular data generators operate on records encoded as sequences of characters, often representing numeric values in fixed formats. From a privacy perspective, these numeric strings constitute a \textit{distinct threat surface}: even minor changes at the character level can correspond to meaningful alterations in sensitive information, such as financial amounts, medical measurements, or personally identifiable metrics. Unlike free-form text, where approximate matches may be semantically ambiguous, we hypothesize that numeric strings are rigid and highly informative, making them particularly vulnerable to memorization by LLMs and thus adversarial signal.  

In this section, we focus on \textit{attacks that exploit these implied strings}. We show that by measuring string similarity between synthetic outputs and potential training records, an adversary can infer membership without access to the model internals, queries, or auxiliary data. This approach exposes a realistic privacy risk inherent in current LLM-based tabular data generation pipelines, emphasizing the need to examine character-level leakage beyond traditional feature-space analyses.

\subsection{Threat Model}

In this work, we adopt a conservative No-box threat model \citep{houssiau2022tapas, Synth-MIA}, in which the adversary has access only to the synthetic data $S$ produced by the model. We make no assumptions that the adversary has knowledge of the model implementation and hyperparameters, query access, or even a reference dataset for constructing calibrated attacks. This threat model is motivated by two key considerations:
\begin{enumerate}
    \item \textbf{Realism}. No-box threat models represent realistic privacy scenarios for tabular synthetic data practitioners. A common scenario involves a user training a tabular data generator on private datasets and sharing the generated data with public or private parties without disclosing implementation details. From the adversary's perspective, this synthetic data may be maliciously acquired or discovered on open data sharing platforms, with no additional context about the generation process.
    \item \textbf{Difficulty}. No-box threat models are recognized as particularly challenging for constructing effective attacks due to the severe limitations placed on the adversary \cite{ganleaks, houssiau2022tapas}. Specifically, the adversary cannot analyze the model's loss function, construct shadow models, or query the model directly. Successful attacks under these restrictive conditions therefore highlight fundamental vulnerabilities in these synthetic data generation methods.
\end{enumerate}

\subsection{Levenshtein Attack}

String similarity is a well-established concept, with Levenshtein edit distance providing a fundamental measure of character-level overlap \citep{levenshtein1966binary, navarro2001guided, Yujian2007}. Recent work has applied normalized variants of this distance to quantify approximate memorization in LLMs, showing that models can reproduce training examples with minor variations in natural language text \citep{ippolito-etal-2023-preventing, shilov2025mosaicmemorylargelanguage}. However, these studies stop short of embedding string similarity into an adversarial framework, as open-ended natural language provides many valid paraphrases and thus yields weak signals for membership inference.

In contrast, \textit{the string representations of values in tabular data are highly constrained.} Unlike free-form text—where two sentences can express identical meaning with entirely different surface forms— tabular records have rigid schemas. Numerical fields follow fixed precision, delimiters appear in consistent positions, categorical fields draw from small vocabularies, and missing values are encoded in standardized ways. As a result, even a single-character modification (e.g., “17.5” → “17.6”) corresponds to a meaningful change. This rigidity turns Levenshtein distance from a fragile signal in natural language into a highly sensitive indicator of memorization in tabular domains.

To operationalize this idea, we convert each record into a canonical string representation that preserves its structure. Numerical features are formatted with consistent precision, categorical variables mapped to stable tokens, and delimiters inserted to mark column boundaries. This step ensures that edit distance reflects true content similarity rather than superficial formatting differences. When a synthetic generator memorizes a training record, its outputs often contain near-exact replicas—matching digit sequences, categorical codes, and formatting patterns. Because such close matches are extremely unlikely to arise by chance, unusually low edit distances between a real record and its nearest synthetic neighbor provide strong evidence of membership.

Motivated by these observations, we design a membership inference attack based on Levenshtein distance. For each test observation $x^*$ and synthetic set $S$, we extract the ordered numeric and categorical values and encode them as strings (see Figure \ref{fig:levattdiagram}), obtaining $x^*{\text{str}}$ and $S_{\text{str}}$. The LevAtt score for Equation \ref{eq:membership_prediction} is defined as:

\begin{equation}
f_{\text{LevAtt}}(x^*_{\text{str}}, S_{\text{str}}) = -\min_{s \in S_{\text{str}}} , \ell(x^*_{\text{str}}, s),
\label{eq:levatt}
\end{equation}

where $\ell$ denotes the Levenshtein edit distance. Lower distances (i.e., higher scores) indicate closer matches and therefore stronger evidence of memorization. 

LevAtt takes as input any type of data including numeric, categorical, missing, and open text, as each of these can be encoded as a string. As LevAtt requires only synthetic outputs, it applies directly to our No-box threat model highlighting a realistic privacy risk in LLM-based tabular data generation. Since the scale of Levenshtein distances varies by dataset, a decision threshold $\gamma$ can be chosen relative to the distribution of scores, such as selecting a value from the Inter-Quartile Range. This approach mirrors the evaluation of DOMIAS \cite{vanbreugel2023membership}, where the median score serves as a threshold to derive binary classification metrics and assess overall attack performance.

\begin{table*}
\centering
\caption{LevAtt performance against ICL models. Mean (STD) values are reported for all datasets, training sizes, and seeds (315 runs Overall) and for the top 20 runs for each model with the highest AUC-ROC (Top 20). GPT-4o-mini shows relatively little privacy leakage, whereas LLama-3.3-70b and TabPFN-V2 show significant privacy failure- especially amongst their worst datasets. At the same time, these models generally have the best fidelity and utility measures.}
\label{tab:icl_results}
\small
\begin{tabular}{ll | ccc | cc | cc}
\toprule
 &  & \multicolumn{3}{c}{LevAtt}
 & \multicolumn{2}{c}{Fidelity}
 & \multicolumn{2}{c}{ML Utility} \\

Run
& Model
 & AUC
 & TPR@0
 & TPR@0.1
 & MMD  $\downarrow$
 & Wass  $\downarrow$
 & AUC  $\uparrow$
 & F1 $\uparrow$\\
\midrule

Overall

& LLaMA-3.3-70B
 & \textbf{0.63 $\pm$ 0.13}
 &\textbf{ 0.08 $\pm$ 0.20}
 & \textbf{0.28 $\pm$ 0.24}
 & \textbf{0.04 $\pm$ 0.02}
 & 1369.27 $\pm$ 22068.62
 & 0.63 $\pm$ 0.12
 & 0.41 $\pm$ 0.18 \\

& TabPFN-V2 
 & 0.58 $\pm$ 0.11
 & 0.04 $\pm$ 0.17
 & 0.19 $\pm$ 0.21
 & 0.04 $\pm$ 0.03
 &\textbf{ 1.01 $\pm$ 1.14}
 & \textbf{0.66 $\pm$ 0.12}
 & \textbf{0.41 $\pm$ 0.18} \\
& GPT-4o-mini
 & 0.54 $\pm$ 0.05
 & 0.00 $\pm$ 0.01
 & 0.13 $\pm$ 0.06
 & 0.05 $\pm$ 0.07
 & 4458.30 $\pm$ 72617.27
 & 0.59 $\pm$ 0.10
 & 0.34 $\pm$ 0.17 \\
\midrule

Top 20

& LLaMA-3.3-70B
 & \textbf{0.97 $\pm$ 0.03}
 & \textbf{0.64 $\pm$ 0.27}
 & 0.93 $\pm$ 0.07
 & \textbf{0.03 $\pm$ 0.02}
 & \textbf{1.38 $\pm$ 1.62}
 & 0.65 $\pm$ 0.11
 & \textbf{0.41 $\pm$ 0.14} \\

& TabPFN-V2
 & 0.97 $\pm$ 0.07
 & 0.57 $\pm$ 0.41
 &\textbf{ 0.94 $\pm$ 0.14}
 & 0.05 $\pm$ 0.05
 & 2.66 $\pm$ 1.62
 &\textbf{ 0.68 $\pm$ 0.16}
 & 0.40 $\pm$ 0.20 \\
& GPT-4o-mini
 & 0.60 $\pm$ 0.09
 & 0.01 $\pm$ 0.03
 & 0.27 $\pm$ 0.09
 & 0.07 $\pm$ 0.05
 & 1.84 $\pm$ 1.80
 & 0.54 $\pm$ 0.05
 & 0.26 $\pm$ 0.08 \\

\bottomrule
\end{tabular}
\end{table*}

% \begin{table*}[t]
% \centering
% \caption{LevAtt performance against ICL models. Mean (STD) values are reported for all datasets, training sizes, and seeds (Overall) and for the top 20 runs for each model with the highest AUC-ROC (Top 20). GPT-4o-mini shows relatively little privacy leakage, whereas LLama-3.3-70b and TabPFN-V2 show significant privacy failure- especially amongst their worst datasets.}
% \label{tab:icl_results}
% \begin{tabular}{l c c c c c c}
% \toprule
% \multirow{2}{*}{Generator} & \multicolumn{2}{c}{AUC-ROC} & \multicolumn{2}{c}{TPR@FPR=0} & \multicolumn{2}{c}{TPR@FPR=0.1} \\
% \cmidrule(lr){2-3} \cmidrule(lr){4-5} \cmidrule(lr){6-7}
%  & Overall & Top 20 & Overall & Top 20 & Overall & Top 20 \\
% \midrule
% LLama-3.3-70b & \textbf{0.63 (0.12)} & \textbf{0.91 (0.03)} & \textbf{0.08 (0.20)} & \textbf{0.64 (0.27)} & \textbf{0.28 (0.21)} & {0.81 (0.09)} \\
% TabPFN-V2     & 0.58 (0.05) & 0.91 (0.07) & 0.04 (0.01) & 0.57 (0.41) & 0.19 (0.06) & \textbf{0.94 (0.14)} \\
% GPT-4o-mini & 0.54 (0.05) & 0.60 (0.09) & 0.01 (0.01) & 0.01 (0.03) & 0.13 (0.05) & 0.27 (0.09) \\
% \bottomrule
% \end{tabular}
% \end{table*}

\section{Experiments}
We evaluate the privacy leakage of string memorization for in-context learning (ICL) and supervised fine-tuning (SFT) LLM-based tabular generators. Here, we use differing experimental designs to correspond to their popular use-case settings. For clarity, we organize the section into three subsections: experimental design details for both regimes and then separate subsections for results. We include the full experimental and implementation details in Appendix: \ref{app:section4}.

\subsection{Experimental Design}
\subsubsection{ICL Approaches}
Replicating the design of \cite{byun2025riskcontextbenchmarkingprivacy}, we evaluate ICL approaches on the OpenML CTR23 benchmark \citep{fischer2023openmlctr}, consisting of 35 real-world regression datasets with 500–100,000 rows and up to 5,000 features, including both numerical and categorical attributes (see Table~\ref{tab:ctr23_metadata}). We partition each dataset into 80/20 training and test sets. To simulate the low-data regime these methods are commonly used for, we subsample ${32, 64, 128}$ training rows without replacement. These sampled rows are provided as exemplars to the ICL models.

We consider three ICL models: LLaMA 3.3 70B \citep{meta2024llama3_3_70b} an open-source foundation model, GPT-4o-mini \citep{openai2024gpt4omini} a close-source foundation model, and TabPFN-v2 \cite{hollmann2025accurate} a transformer-based model trained on tabular data due to their wide use and availability. For each sampled training subset, we generate an equal amount of synthetic data. Evaluation is conducted on all exemplar rows plus an equal holdout partition of the test set. We evaluate privacy leakage using the following No-box MIAs using the Synth-MIA package \citep{Synth-MIA}: LevAtt, Euclidean Distance to Closest Record (DCR) \citep{ganleaks}, a Monte Carlo density estimation (MC) method \citep{Hilprecht2019MonteCA}, and a kernel density estimation method \citep{houssiau2022tapas, vanbreugel2023membership}. We report MIA success using mean AUC-ROC and TPR at fixed FPR thresholds \citep{Carlini2021MembershipIA} to capture potential information leakage. We repeat this experimentation across 3 seeds, totaling 315 datasets/ subsample sizes/ seeds per model.

\subsubsection{SFT Approaches}
\label{SFT Experiments}
For SFT-approaches, we benchmark the original SFT-based tabular generation method GREAT \cite{borisov2023languagemodelsrealistictabular} and RealTabFormer \citep{realtabformer}, which reports improved privacy performance due to enforcing a minimum Euclidean Distance to Closest to Record distribution in its training. As both of these methods use GPT-2 \citep{radford2019language}as a base model, we modify RealTabFormer to accept more modern, larger foundation models: LLaMA 3.2 (1B, 3B) \citep{grattafiori2024llama3herdmodels}, Qwen2.5-3B \citep{qwen2025qwen25technicalreport}, and Mistral v0.3 7B \cite{jiang2023mistral7b}. Additionally, we introduce as a control CT-GAN and TVAE \cite{Xu2019ModelingTD}, conventional deep learning-based generators. Lastly, we benchmark DP2Stage \cite{afonja2025dpstage}, a differentially private framework that SFT's GPT-2 with DP-SGD. Training follows default hyperparameters from the original GREAT,  RealTabFormer, and DP2Stage implementations while CTGAN and TVAE are implemented through Synthcity \cite{synthcity}.

Experiments are conducted on five tabular datasets: CASP, Aba\-lone, Diabetes, CA-Housing, and Faults (see Table~\ref{tab:sft_datasets}), selected for their common use in synthetic tabular data benchmarking and containing numeric data. Similarly to ICL, we create 80/20 train-test partitions and following common synthetic tabular data benchmarking \citep{tabsyn}, synthetic datasets equal in size to the original training sets are generated post-training. For privacy evaluation, up to 1,000 training and holdout samples are partitioned as test sets, and the same MIAs are applied. Again, we repeat this experimentation across 3 seeded runs.
\begin{figure}[t!]
    \centering
    \includegraphics[width=.75\linewidth]{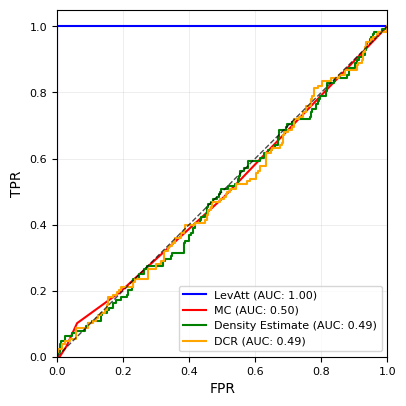}
    \caption{ROC plot for various No-box MIAs against TabPFN-V2 with 128 in-context samples from the MoneyBall dataset. LevAtt (blue) is able to achieve perfect classification for all in-context samples whereas MIAs that target the feature space of tabular data fail to capture the privacy leakage.}
    \Description{ROC curve plot comparing multiple membership inference attacks. The LevAtt method (blue line) achieves perfect classification with AUC of 1.0, reaching the top-left corner, while other methods (DCR, Density Estimate, and MC) perform poorly near the diagonal baseline.}
    \label{fig:roc_curve_tabpfn}
\end{figure}

\begin{figure}
    \centering
    \includegraphics[width=.75\linewidth]{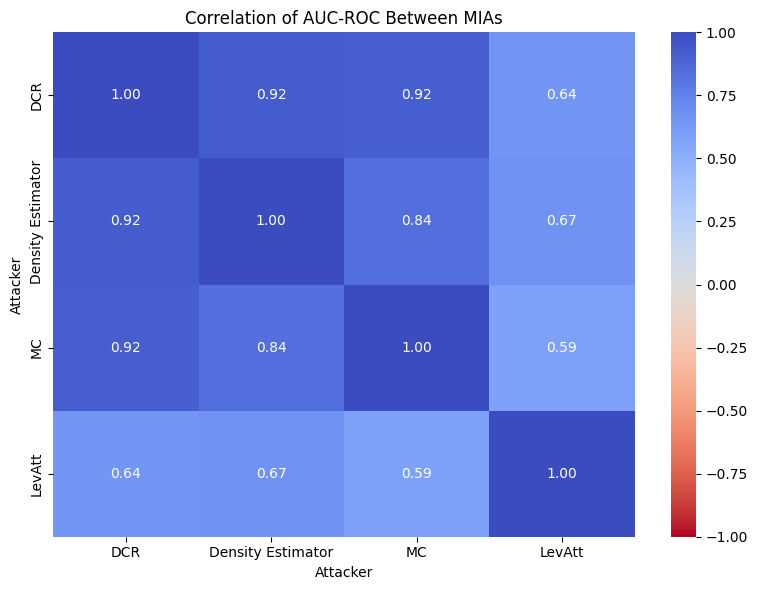}
    \caption{Correlation plot for No-box MIA AUC-ROC across the ICL experiment. While the feature-space targeting DCR, Density Estimate, and MC are nearly perfectly correlated, LevAtt is much less correlated. This highlights that while privacy leakage over tabular string representations and the feature space are related, LevAtt finds unique adversarial advantage.}
    \Description{Correlation heatmap showing AUC-ROC correlations between different MIA methods. DCR, Density Estimate, and MC show near-perfect correlation (values close to 1.0), while LevAtt shows notably lower correlation with these methods, indicating it captures different privacy leakage patterns.}
    \label{fig:icl_cor_plot}
\end{figure}
\subsection{Metrics}

We evaluate the synthetic data generators across three primary dimensions: privacy, fidelity, and utility. To quantify privacy, we measure the success of LevAtt using the Area Under the ROC Curve (AUC) and the True Positive Rate (TPR) at low False Positive Rate (FPR) thresholds. Following \citet{Carlini2021MembershipIA}, we prioritize these non-thresholded metrics because fixed-threshold accuracy can mask score separability, thereby underestimating the risk of memorization. Fidelity is assessed by calculating the Maximum Mean Discrepancy (MMD) with an RBF kernel and the Wasserstein distance between the real and synthetic distributions, where lower values signify superior distributional alignment. Finally, we evaluate utility via a "Train on Synthetic, Test on Real" (TSTR) framework \cite{synthcity}; we train an XGBoost classifier on the synthetic data and report the AUC and F1 score on a held-out real dataset to determine how effectively the synthetic samples preserve task-relevant structure.

\subsection{Results: ICL Methods}

Our ICL experiments reveal clear evidence that LLM-based tabular generators can exhibit substantial privacy leakage, even under an extremely restrictive threat model. We organize our findings around three central observations.

\textbf{LevAtt identifies substantial privacy leakage in ICL-based tabular generators.} 
Table~\ref{tab:icl_results} reports the Mean (STD) of LevAtt AUC-ROC and TPR@FPR $\in \{0.0, 0.1\}$ across all datasets, training sizes, and seeds. Because privacy audits often emphasize worst-case behavior, we also report metrics for the top 20 runs (ranked by AUC-ROC) for each model. This analysis reveals that even a simple string-similarity attack can be highly effective against state-of-the-art ICL generators. LLama~3.3-70B and TabPFN-V2 are particularly vulnerable, with mean TPR@FPR=0 values of 0.64 and 0.57 respectively in their highest-performing runs. Strikingly, LevAtt even achieves \emph{perfect} membership classification in several TabPFN-V2 settings (Figure~\ref{fig:roc_curve_tabpfn}), demonstrating that memorized digit sequences can sometimes be reproduced verbatim in generated samples. 

Additionally, we analyze the privacy leakage of each model based on the number of in-context examples in Table~\ref{tab:icl_n_size} (Appendix~\ref{sec:tables}). Here, we find that for all models, runs with fewer examples consistently experienced greater privacy leakage. This indicates that at small sample sizes, ICL data generation strategies have a greater tendency to directly memorize the string representations of example data.

\textbf{Privacy leakage scales with model size.}
These results are consistent with prior evidence that memorization grows with model capacity~\citep{carlini2021extractingtrainingdatalarge, carlini2023quantifyingmemorizationneurallanguage}. Although the parameter counts of GPT-4o-mini and TabPFN-V2 are not publicly disclosed, empirical benchmarks place GPT-4o-mini near the 7B scale while TabPFN-V2 is lightweight enough for single-GPU inference. In contrast, the 70B-parameter LLaMA~3.3 shows the largest privacy leakage across our benchmark. This pattern reinforces that larger models, even in an ICL setting without fine-tuning, have greater capacity to memorize and regenerate specific training examples.

\textbf{LevAtt captures a distinct leakage signal relative to feature-space MIAs.}
To evaluate whether LevAtt overlaps with traditional feature-space MIAs, we analyze its correlation with DCR, Density Estimation, and MC attacks (Figure~\ref{fig:icl_cor_plot}). The three feature-space attacks are nearly perfectly correlated with each other, reflecting their shared reliance on density discrepancies. In contrast, LevAtt exhibits substantially weaker correlation with all three, indicating that it identifies leakage signals orthogonal to feature-space methods. Indeed, in some instances, we found that LevAtt was a perfect classifier of membership in some experimental runs whereas traditional feature-space MIAs with compatible threat models were no better than random. This highlights that LLM-based tabular generators introduce a novel, string-level memorization vector that would be entirely missed by conventional synthetic-data MIAs.
% This distinction is particularly important as recent work has demonstrated that even correlated attacks can be ensembled to create more powerful classifiers \citep{Ensemble MIA}.
\subsection{Results: SFT Methods}
\label{sub: results sft}
\begin{table*}
\caption{(Mean $\pm$ STD) LevAtt performance for each model and dataset across seeds. RealTabFormer experiences significant privacy leakage and LevAtt finds some signal for most LLM-based models in various datasets. However, LevAtt identifies no leakage in conventional deep learning-based methods CT-GAN and TVAE as they do not generate strings. (- indicates the model did not converge.)}
\label{tab:sft-levatt}
\begin{tabular}{llccccc}
\toprule
Model & Metric & Housing & Diabetes & CASP & Faults & Abalone \\
\midrule
\multirow[t]{2}{*}{RealTabFormer} & Lev-AUC & \bm{$0.62 \pm 0.01$} & \bm{$0.67 \pm 0.01$} & \bm{$0.70 \pm 0.01$} &\bm{ $0.69 \pm 0.00$} & \bm{$0.59 \pm 0.00$} \\
 & TPR@FPR=0.1 & \bm{$0.17 \pm 0.02$} & \underline{$0.25 \pm 0.02$} & \bm{$0.35 \pm 0.02$} & \bm{$0.33 \pm 0.01$} & \bm{$0.17 \pm 0.00$} \\
\cline{1-7}
\multirow[t]{2}{*}{LLaMA 3.2-1B} & Lev-AUC & $0.58 \pm 0.00$ & $0.57 \pm 0.00$ & \underline{$0.51 \pm 0.01$} & \underline{$0.56 \pm 0.01$} & $0.52 \pm 0.00$ \\
 & TPR@FPR=0.1 & $0.13 \pm 0.00$ & $0.21 \pm 0.00$ & $0.09 \pm 0.01$ & \underline{$0.16 \pm 0.01$} & $0.12 \pm 0.01$ \\
\cline{1-7}
\multirow[t]{2}{*}{LLaMA 3.2-3B} & Lev-AUC & \underline{$0.59 \pm 0.01$} & $0.57 \pm 0.01$ & $0.50 \pm 0.00$ & $0.55 \pm 0.01$ & \underline{$0.57 \pm 0.00$} \\
 & TPR@FPR=0.1 & $0.14 \pm 0.01$ & $0.17 \pm 0.02$ & $0.09 \pm 0.01$ & $0.13 \pm 0.01$ & \underline{$0.17 \pm 0.01$} \\
\cline{1-7}
\multirow[t]{2}{*}{GPT2} & Lev-AUC & $0.59 \pm 0.00$ & $0.51 \pm 0.00$ & $0.51 \pm 0.01$ & $0.52 \pm 0.00$ & --- \\
 & TPR@FPR=0.1 & \underline{$0.14 \pm 0.01$} & $0.09 \pm 0.01$ & \underline{$0.10 \pm 0.00$} & $0.14 \pm 0.01$ & --- \\
\cline{1-7}
\multirow[t]{2}{*}{GReaT} & Lev-AUC & $0.58 \pm 0.01$ & $0.50 \pm 0.01$ & $0.52 \pm 0.00$ & --- & $0.51 \pm 0.00$ \\
 & TPR@FPR=0.1 & $0.12 \pm 0.02$ & $0.09 \pm 0.00$ & $0.10 \pm 0.00$ & --- & $0.11 \pm 0.01$ \\
\cline{1-7}
\multirow[t]{2}{*}{Qwen 2.5-3B} & Lev-AUC & $0.59 \pm 0.00$ &\underline{$0.64 \pm 0.00$} & $0.51 \pm 0.01$ & $0.54 \pm 0.01$ & $0.51 \pm 0.01$ \\
 & TPR@FPR=0.1 & $0.13 \pm 0.01$ & \bm{$0.27 \pm 0.02$} & $0.10 \pm 0.01$ & $0.11 \pm 0.01$ & $0.11 \pm 0.01$ \\
\cline{1-7}
\multirow[t]{2}{*}{Mistral-7B} & Lev-AUC & $0.58 \pm 0.00$ & $0.52 \pm 0.00$ & $0.51 \pm 0.02$ & $0.51 \pm 0.00$ & $0.50 \pm 0.00$ \\
 & TPR@FPR=0.1 & $0.13 \pm 0.01$ & $0.10 \pm 0.00$ & $0.10 \pm 0.02$ & $0.11 \pm 0.01$ & $0.11 \pm 0.01$ \\
\cline{1-7}
\multirow[t]{2}{*}{TVAE} & Lev-AUC & $0.51 \pm 0.00$ & $0.50 \pm 0.01$ & $0.48 \pm 0.00$ & $0.50 \pm 0.00$ & $0.49 \pm 0.01$ \\
 & TPR@FPR=0.1 & $0.09 \pm 0.00$ & $0.10 \pm 0.01$ & $0.10 \pm 0.00$ & $0.08 \pm 0.01$ & $0.08 \pm 0.01$ \\
\cline{1-7}
\multirow[t]{2}{*}{CT-GAN} & Lev-AUC & $0.50 \pm 0.01$ & $0.48 \pm 0.00$ & $0.49 \pm 0.01$ & $0.49 \pm 0.01$ & $0.50 \pm 0.00$ \\
 & TPR@FPR=0.1 & $0.09 \pm 0.01$ & $0.09 \pm 0.00$ & $0.10 \pm 0.01$ & $0.09 \pm 0.02$ & $0.06 \pm 0.00$ \\
\cline{1-7}
\multirow[t]{2}{*}{DP2Stage  ($\epsilon=10$)} & Lev-AUC & $0.53 \pm 0.00$ & $0.50 \pm 0.01$ & $0.48 \pm 0.00$ & --- & $0.48 \pm 0.00$ \\
 & TPR@FPR=0.1 & $0.10 \pm 0.01$ & $0.11 \pm 0.02$ & $0.09 \pm 0.01$ & --- & $0.08 \pm 0.01$ \\
\cline{1-7}
\bottomrule
\end{tabular}
\end{table*}

We now turn to the SFT regime, where models are explicitly trained to approximate the full joint distribution of the training data. While leakage is generally less severe than in ICL, our findings demonstrate that SFT does not eliminate privacy risk and that leakage systematically depends on model design, sampling volume, and data structure.

\textbf{SFT approaches exhibit noticeable but lower privacy leakage.}
Table~\ref{tab:sft-levatt} summarizes LevAtt performance across all datasets and seeds. Compared to ICL, SFT models typically show reduced leakage; however, RealTabFormer consistently exhibits the highest vulnerability—despite incorporating privacy-aware training—and does so more prominently than GREAT, despite their shared GPT-2 base. LevAtt also detects measurable leakage in LLaMA-3.2 (1B and 3B) and Qwen-2.5-3B. By contrast, LevAtt is ineffective against CT-GAN and TVAE, which generate full tabular rows rather than token sequences. This reinforces that token-level generation provides an attack surface absent in conventional deep-learning tabular synthesizers.

\begin{figure}
% Figure part
    \centering
    \includegraphics[width=\linewidth]{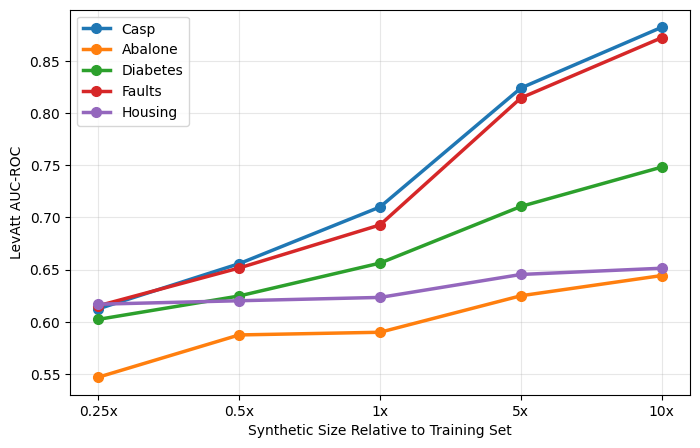}
    \caption{LevAtt AUC-ROC for various datasets generated by RealTabFormer with increasing synthetic dataset sizes relative to the training set.}
    \Description{Line plot showing LevAtt AUC-ROC performance across multiple datasets as synthetic dataset size increases from 1x to 10x the training set size. Multiple colored lines represent different datasets, generally showing increasing AUC-ROC values as synthetic data size increases.}
    \label{fig:synth_size}
\end{figure}

\textbf{Privacy leakage increases with the volume of synthetic data.}
To understand how synthetic data scale affects leakage, we generate datasets at 0.25$\times$, 0.5$\times$. 1$\times$, 5$\times$, and 10$\times$ the size of the original training data using RealTabFormer. As shown in Figure~\ref{fig:synth_size}, LevAtt’s AUC-ROC increases monotonically with synthetic dataset size, with some datasets (e.g., Faults) exhibiting up to a 20\% absolute increase from 1$\times$ to 10$\times$. At lower levels, 0.25$\times$ and 0.5$\times$, LevAtt sees worse performance as there is less information released from the generator. This suggests a simple mechanism: emitting more samples increases the likelihood that a memorized training example is reproduced. This has direct implications for practitioners who rely on synthetic data to replace or augment sensitive datasets—larger synthetic releases can inadvertently amplify privacy risk.

\begin{figure}
    \centering
    \includegraphics[width=\linewidth]{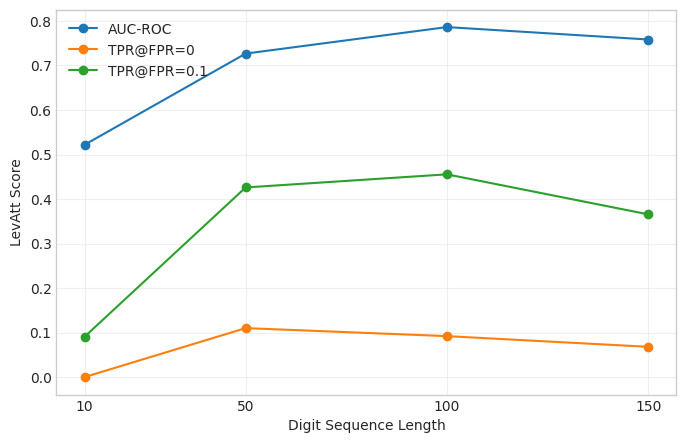}
    \caption{LevAtt performance on RealTabFormer at various training digit sequence lengths.}
    \Description{Plot showing the relationship between digit sequence length in training data and LevAtt attack performance on RealTabFormer. The plot demonstrates how AUC-ROC increases with different lengths of digit sequences used during training.}
    \label{fig:digit_count}
\end{figure}

\textbf{Memorization increases with sequence length in the training data.}
We study how digit sequence length affects memorization. Here, we create training and holdout sets using identical multivariate Gaussian distributions with a controlled numbers of digit columns. We then vary the total sequence length of digits for a row while keeping these distribution fixed. After training RealTabFormer on datasets with progressively longer sequences, we observe that LevAtt performance increases accordingly (Figure~\ref{fig:digit_count}) before leveling off at 100 digits. This replicates earlier findings that longer sequences encourage LLM memorization~\citep{carlini2021extractingtrainingdatalarge} and highlights that dataset structure—not only model architecture—plays a direct role in privacy risk. Datasets containing long numeric strings (e.g., identifiers, timestamps, financial fields) inherently create more opportunities for memorization and unintended reproduction.

\textbf{LevAtt Privacy vs Fidelity Tradeoff.}
Finally, the empirical results in Table \ref{tab:lev_fidelity_results} reveal a tension between generative utility and LevAtt susceptibility. Most notably, RealTabFormer achieves state-of-the-art performance ($0.92 \pm 0.11$ ML AUC) but incurs the highest empirical privacy risk ($0.65 \pm 0.04$ Lev AUC), suggesting that high-fidelity synthesis often correlates with increased record-level leakage. This trade-off is further quantified by the negative correlation between Lev AUC and statistical distance metrics like MMD ($-0.163$) and Wasserstein distance ($-0.084$), indicating that as LLMs minimize distributional divergence, they become more vulnerable to string similarity attacks. We also observe that incorporating larger, more modern language models into the RealTabFormer framework does not yield improvements in synthetic data quality. We hypothesize that this performance gap stems from a mismatch between model architecture and hyperparameter configuration: the original models were carefully tuned for the framework, whereas the newer models likely require different optimization settings to fully realize their capacity.

% \begin{table}[ht]
% \small
% \centering
% \caption{Lev-MIA Results for real-world datasets and synthetic data generated by Realtabformer}

% \begin{tabular}{llccccc}

% \hline
% \textbf{Dataset} & \textbf{Synth Size} & \textbf{AUC ROC} & \textbf{TPR@FPR=0} & \textbf{TPR@FPR=0.001} & \textbf{TPR@FPR=0.01} & \textbf{TPR@FPR=0.1} \\
% \hline
% CASP & 10x & 0.88 & 0.01 & 0.01 & 0.10 & 0.67 \\
% CASP & 1x  & 0.71 & 0.00 & 0.00 & 0.04 & 0.37 \\
% abalone & 10x & 0.57 & 0.00 & 0.00 & 0.02 & 0.16 \\
% abalone & 1x  & 0.55 & 0.01 & 0.01 & 0.02 & 0.14 \\
% kc2 & 10x & 0.79 & 0.00 & 0.00 & 0.04 & 0.43 \\
% kc2 & 1x  & 0.63 & 0.00 & 0.00 & 0.03 & 0.23 \\
% diabetes & 10x & 0.67 & 0.01 & 0.02 & 0.04 & 0.22 \\
% diabetes & 1x  & 0.63 & 0.01 & 0.01 & 0.04 & 0.21 \\
% housing & 10x & 0.65 & 0.00 & 0.00 & 0.03 & 0.19 \\
% housing & 1x  & 0.64 & 0.00 & 0.00 & 0.02 & 0.19 \\
% \hline
% \end{tabular}
% \label{tab:lev_metrics}
% \end{table}

\section{Defenses Against Levenshtein Attack}

The efficacy of LevAtt against LLM-based tabular models motivates us to explore methodologies that can defend generated synthetic data. Here, recognizing that LevAtt gains adversarial advantage from replicated patterns in strings of digits, we devise two strategies based on introducing controlled "noise" into string sequences. The first is Digit Modifier (DM), a post-processing method that operates independently of the generative model and is applied after the synthetic data have been produced. The second protection strategy is Tendency-based Logit Processor (TLP), a method compatible with any open-source LLM that strategically perturbs digits at sample time.

% In this section, we introduce our defense mechanism, the tendency-based logit processor. Specifically, in section \ref{sub: DM and TLP} we will first introduce the most intuitive way of defense: the digit modifier along with our tendency logit processor. We provide intuition for both protection mechanisms, explaining why they should be effective. Then, in section \ref{sub:limits}, we will argue, along with simulation, to demonstrate why the digit modifier is not a robust mechanism. Afterwards, in section \ref{sub:Efficacy of TLP} we will show that our tendency-based logit processor effectively reduces detection performance of the adversary attacks, while maintaining relatively good fidelity and utility score. In particular, all experiments in this section were conducted with RealTabFormer, which employs GPT-2 as its base LLM \citep{realtabformer}.
\begin{table}[h!]
\footnotesize
\caption{Model Privacy and Fidelity Tradeoffs. We report the (Mean $\pm$ STD) of LevAtt's AUC, Maximum Mean Discrepency, Wasserstein Distance, and Downstream ML AUC for each model over all datsets as well as the Pearson Correlation of all metrics over all runs with LevAtt AUC.}
\begin{tabular}{lcccc}
\hline
{Model} & {Lev AUC} & {MMD $\downarrow$} & {Wass $\downarrow$} & {ML AUC $\uparrow$} \\
\hline
RealTabFormer & \bm{$0.65 \pm 0.04$} & $0.01 \pm 0.00$ &\bm{ $0.13 \pm 0.17$} & \bm{$0.92 \pm 0.11$} \\
LLaMA-1B & $0.54 \pm 0.03$ & $0.03 \pm 0.03$ & $0.31 \pm 0.34$ & \underline{$0.89 \pm 0.14$} \\
LLaMA-3.2-3B & \underline{$0.56 \pm 0.03$} & $0.02 \pm 0.02$ & $0.33 \pm 0.48$ & $0.83 \pm 0.12$ \\

GPT-2 & $0.53 \pm 0.04$ & $0.01 \pm 0.00$ & $0.45 \pm 0.65$ & $0.75 \pm 0.19$ \\
GReaT & $0.53 \pm 0.03$ & $0.01 \pm 0.00$ & $7.00 \pm 23.44$ & $0.83 \pm 0.26$ \\
Mistral & $0.53 \pm 0.03$ & $0.02 \pm 0.01$ & $16.40 \pm 33.60$ & $0.64 \pm 0.16$ \\
Qwen-3B & $0.56 \pm 0.05$ & $0.01 \pm 0.00$ & $0.20 \pm 0.27$ & $0.88 \pm 0.16$ \\
TVAE & $0.50 \pm 0.01$ & $0.01 \pm 0.00$ & \underline{$0.19 \pm 0.26$} & $0.88 \pm 0.13$ \\
CTGAN & $0.49 \pm 0.01$ & $0.01 \pm 0.00$ & $0.24 \pm 0.34$ & $0.85 \pm 0.09$ \\
DP-2Stage & $0.50 \pm 0.02$ & $0.04 \pm 0.06$ & $1.08 \pm 0.57$ & $0.50 \pm 0.01$ \\
\hline
{Corr. Lev AUC} & --- & $-0.163$ & $-0.084$ & $0.388$ \\
\hline
\end{tabular}
\label{tab:lev_fidelity_results}
\end{table}
\subsection{Digit Modifier}
\label{sub: DM and TLP}
We propose a Digit Modifier (DM), a method for injecting controlled randomness into tabular data by perturbing numerical digits. Motivated by bit-flipping techniques for adding noise to binary representations of relational databases \cite{agrawal2002watermarking,alfagi2016survey}, DM operates by replacing tokens in a record’s tokenized sequence with alternatives sampled from a noise distribution. When tokens represent numerical values, these substitutions correspondingly alter the digits of the underlying sequences, yielding perturbed records. In this sense, DM can be viewed as a principled method for randomly replacing digits within a record.

To balance data fidelity and protection, we parameterize the mechanism as $\text{DM}(p_{min}, p_{max})$ with $0 \leq p_{min} < p_{max} \leq 1$. For a numerical column $\mathbf{X_i}$ and entry $\mathbf{x_{k,i}}$, a probability function $g: \mathbf{x_{k,i}} \times \mathbf{X_i} \to [p_{min}, p_{max}]$ assigns digit-flipping probabilities such that larger-magnitude entries receive higher perturbation probabilities. Each digit of $\mathbf{x_{k,i}}$ is then independently flipped according to $g(\mathbf{x_{k,i}}, \mathbf{X_i})$, while smaller-magnitude values—being more sensitive—undergo smaller changes to preserve fidelity. In our experiment, we first define
 
\[
M(\mathbf{X_i}) = \max_{\mathbf{x_{k,i}} \in \mathbf{X_i}} |\mathbf{x_{k,i}}|,
\]
which represents the largest absolute value in the numerical column. We can then constructed the  probability function by
\[
g(\mathbf{x_{k,i}}, \mathbf{X_i}; p_{\min}, p_{\max}) \;=\; p_{\min} + (p_{\max} - p_{\min}) \frac{|\mathbf{x_{k,i}}|}{M(\mathbf{X_i})}.
\]

At a high level, DM operates by iterating through each numerical entry in a tabular record, computing its digit-flipping probability using the function \( g \), and independently perturbing each digit according to that probability. This produces a randomized yet fidelity-preserving transformation in which large-magnitude values receive stronger perturbations while small values remain close to their originals. We summarize the overall DM procedure in Algorithm \ref{algorithm: DM}.

\subsection{Tendency-Based Logit Processor}
We additionally propose Tendency-based Logit Processor ($\text{TLP}$), a mechanism for injecting controlled noise into synthetic data by perturbing the generator’s logits at inference time. The $\text{TLP}(t)$ selectively amplifies lower-valued logits while suppressing higher-valued ones, making the generator more likely to select tokens that were originally less probable. The strength of this effect is controlled by the tendency parameter $t$: higher values of $t$ induce stronger curvature, increasing the randomness of the generated sequence. In this way, $\text{TLP}(t)$ acts as a principled method for introducing variability into synthetic outputs while still preserving the generator’s learned distribution.

Formally, $\text{TLP}(t)$ first maps the generator’s logits $l = (l_1, l_2, ., l_k)$ into the range $[0,1]$ using a shifted min--max scaler $S_l$. Specifically, let $m_l = \min_j l_j$, $M_l = \max_j l_j$, and $\varepsilon > 0$ (small constant). Then, we define 
\[
\big[S_l(l_1, l_2, \ldots, l_n)\big]_i
= \frac{l_i - m_l}{M_l - m_l + \varepsilon}, \qquad i = 1, \ldots, k.
\]
Similarly, we can also define $S^{-1}_l$ componentwise as
\[
\big[S^{-1}_l(s_1, s_2, \ldots, s_n)\big]_i 
= m_l + s_i \,(M_l - m_l + \varepsilon), 
\qquad i = 1, \ldots, n,
\]
which maps processed logits back to their original scale. 
\begin{algorithm}[t]
    \caption{Digit Modifier (DM)}
    \label{algorithm: DM}
    \begin{algorithmic}[1]
        \State \textbf{Input:} Training dataset $D = \{\mathbf{x}_k\}_{k=1}^n$, generator $G$, parameters $0 \leq p_{\min} < p_{\max} \leq 1$, probability function $g$
        \State \textbf{Output:} Perturbed synthetic dataset $\tilde{D}_{\text{syn}}$
        \State Train generator $G$ on dataset $D$
        \State Generate synthetic dataset $D_{\text{syn}} = \{\mathbf{x}_k\}_{k=1}^{n'}$ using $G$
        \State Let $\mathcal{N} \subseteq \{1, \ldots, d\}$ denote the indices of numerical columns
        \ForAll{$i \in \mathcal{N}$}
            \State $\mathbf{X}_i \gets \{x_{1,i}, x_{2,i}, \ldots, x_{n',i}\}$ \Comment{Column $i$ values}
            \State $M(\mathbf{X}_i) \gets \max_{\mathbf{x}_{k,i} \in \mathbf{X}_i} |\mathbf{x}_{k,i}|$ \Comment{Maximum absolute value}
        \EndFor
        \State Initialize $\tilde{D}_{\text{syn}} \gets D_{\text{syn}}$
        \ForAll{$k \in \{1, \ldots, n'\}$} \Comment{For each record}
            \ForAll{$i \in \mathcal{N}$} \Comment{For each numerical column}
                \State $p \gets g(\mathbf{x}_{k,i}, \mathbf{X}_i; p_{\min}, p_{\max})$ \Comment{Compute flip probability}
                \State Let $\mathcal{R}(\mathbf{x}_{k,i})$ denote the set of digit positions in $\mathbf{x}_{k,i}$
                \ForAll{$r \in \mathcal{R}(\mathbf{x}_{k,i})$} \Comment{For each digit position}
                    \State Sample $b \sim \text{Bernoulli}(p)$
                    \If{$b = 1$}
                        \State Let $d_r$ be the digit at position $r$ in $\mathbf{x}_{k,i}$
                        \State Sample $d'_r \sim \text{Uniform}(\{0, 1, \ldots, 9\} \setminus \{d_r\})$
                        \State Replace digit at position $r$ in $\tilde{\mathbf{x}}_{k,i}$ with $d'_r$
                    \EndIf
                \EndFor
            \EndFor
        \EndFor
        \State \textbf{return} $\tilde{D}_{\text{syn}}$
    \end{algorithmic}
\end{algorithm}
Next, TLP applies a monotone increasing, concave curving function $f_t:[0,1] \to [0,1]$, parameterized by $t$ and satisfying $f_t(0)=0$, to each scaled logit. Finally, the processed logits are transformed back to their original scale using the inverse scaler $S_l^{-1}$. The design of $f_t$ is central to $\text{TLP}(t)$. Monotonicity ensures that the relative order of logits is preserved, so high-probability tokens remain more likely than lower-probability ones, allowing controlled noise injection without overwhelming the generator’s learned signal. Concavity, combined with $f_t(0)=0$, guarantees that lower logits are curved upward, increasing their chance of being sampled. Too see why we need concavity, let's fix $0 < a < b \leq 1$. Since $a = \theta b$ for some $\theta \in (0,1)$, concavity of $f_t$ implies
\[
f_t(a) = f_t(\theta b + (1-\theta)\cdot 0) 
\;\geq\; \theta f_t(b) + (1-\theta) f_t(0) 
= \theta f_t(b),
\]
where we used $f(0)=0$. Dividing both sides by $a=\theta b$ gives
\[
\frac{f_t(a)}{a} \;\geq\; \frac{f_t(b)}{b}.
\]
Thus, the ratio $f_t(x)/x$ is non-increasing in $x$. 
This means smaller logits receive a proportionally larger boost compared to larger logits. 
Therefore, $f_t$ preserves the ordering of the logits (by monotonicity) 
while compressing their differences (by concavity), 
which corresponds to a ``curving-up'' transformation that favors lower logits.

\begin{figure*}[ht]
    % \label{fig: Logit Distribution Transformation}
    \centering
    \includegraphics[width=0.75\textwidth]{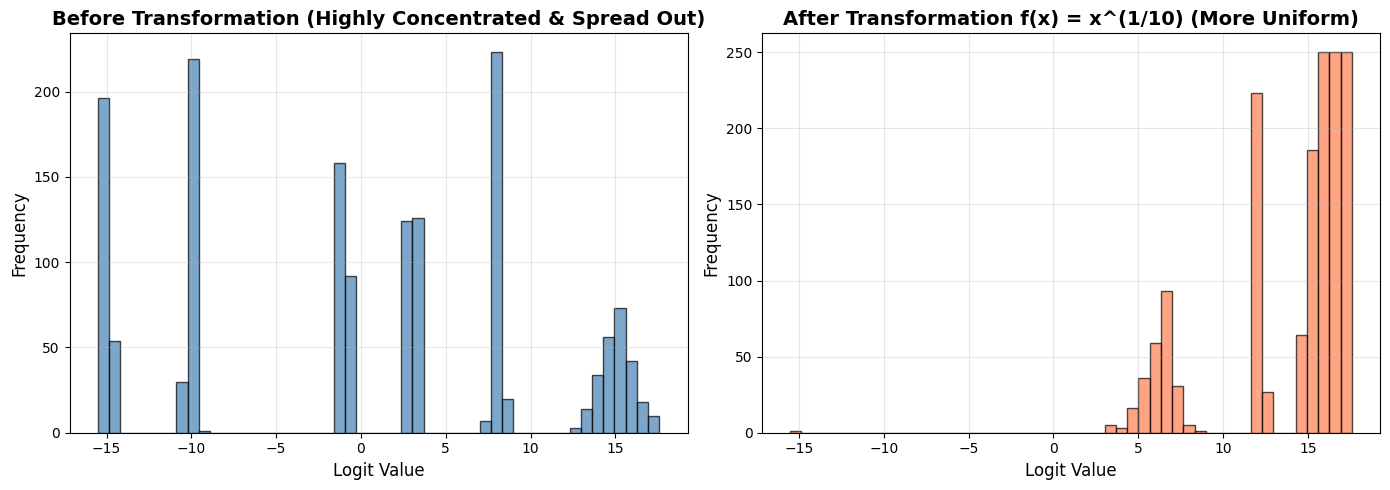}
    \caption{Effect of the TLP transformation on logit distributions. Before transformation (left), lower logits are tightly concentrated near the bottom of the range and have little chance of being selected. After applying the TLP function $f(x) = x^{1/10}$ (right), the transformation increases their relative magnitude of lower value logits, making them more likely to be sampled.}
    \Description{Side-by-side comparison of logit distributions before and after TLP transformation. The left panel shows the original distribution with lower logits compressed at the bottom. The right panel shows the transformed distribution where lower logits are spread out and elevated, demonstrating increased sampling probability for previously low-value logits.}
    \label{fig: Logit Distribution Transformation}
\end{figure*}

In particular, the curving function we use in our experiment is $f_t(x) = x^{\frac{1}{t}}$. Figure \ref{fig:power_roots} below demonstrates the graph of our $f_t$ under different $t$. Figure \ref{fig: Logit Distribution Transformation} shows how $f_t$ transform the distribution of logits so that lower logits get higher probability of being selected. We summarize the overall procedure of TLP in Algorithm \ref{algorithm: TLP}. Please refer to appendix \ref{sub:More Details On Defenses} for more details on DM and TLP.

% TLP_visualization.png
% Concavity visualization of f_t(x) = x^{1/t}
\begin{algorithm}[t]
    \caption{Tendency-Based Logit Processor (TLP)}
    \label{algorithm: TLP}
    \begin{algorithmic}[1]
        \State \textbf{Input:} Training dataset $D$, generator $G$, tendency parameter $t > 0$, curving function $f_t:[0,1]\to[0,1]$
        \State \textbf{Output:} Synthetic sequence generated by $G$ with TLP applied
        \State Train generator $G$ on dataset $D$
        \State Initialize output sequence $\mathbf{y} \gets \emptyset$
        \While{generation not complete}
            \State Compute raw logits $\mathbf{l} = (l_1, \ldots, l_k) \gets G(\mathbf{y})$ \Comment{$k$ = vocabulary size}
            \State $m_l \gets \min_{j \in [k]} l_j$, \quad $M_l \gets \max_{j \in [k]} l_j$
            \State $\mathbf{s} \gets S_l(\mathbf{l})$ where $s_i = \frac{l_i - m_l}{M_l - m_l + \varepsilon}$ for $i \in [k]$ \Comment{Scale to $[0,1]$}
            \State $\tilde{\mathbf{s}} \gets f_t(\mathbf{s})$ where $\tilde{s}_i = f_t(s_i)$ for $i \in [k]$ \Comment{Apply curving function}
            \State $\tilde{\mathbf{l}} \gets S_l^{-1}(\tilde{\mathbf{s}})$ where $\tilde{l}_i = m_l + \tilde{s}_i(M_l - m_l + \varepsilon)$ for $i \in [k]$ \Comment{Rescale}
            \State $\mathbf{p} \gets \text{softmax}(\tilde{\mathbf{l}})$ where $p_i = \frac{\exp(\tilde{l}_i)}{\sum_{j=1}^k \exp(\tilde{l}_j)}$ 
            \State Sample token $y_{\text{next}} \sim \text{Categorical}(\mathbf{p})$
            \State $\mathbf{y} \gets \mathbf{y} \cup \{y_{\text{next}}\}$
        \EndWhile
        \State \textbf{return} $\mathbf{y}$
    \end{algorithmic}
\end{algorithm}

\subsection{Evaluating Defense Mechanisms}
\label{sub:limits}
To evaluate the effectiveness of DM and $\text{TLP}$, we use RealTabFormer, the SFT model exhibiting the highest degree of privacy leakage from LevAtt, and measure how much privacy improvement each method provides against LevAtt while preserving synthetic data fidelity. Here, we use two high privacy leakage datasets for RealTabFormer: a simulated dataset from a Multivariate Gaussian N(300, 5) designed to have 100 digits, and the CASP dataset. We copy the experiment design of Section \ref{SFT Experiments} and sample models at varying synthetic set sizes to induce different levels of privacy leakage in the synthetic datasets.

We then apply DM and $\text{TLP}$ across different scaler and tendency parameter levels respectively, and assess performance in terms of privacy (LevAtt AUC-ROC and TPR@FPR=0.1), fidelity (Wasserstein Distance and Maximum Mean Discrepancy (MMD) between training and synthetic datasets), and utility (RMSE of XGBoost models trained on synthetic data and evaluated on real holdout data) \citep{synthcity}.

\begin{figure}[t]
    \centering
    \begin{subfigure}[t]{0.22\textwidth}
        \includegraphics[width=\linewidth]{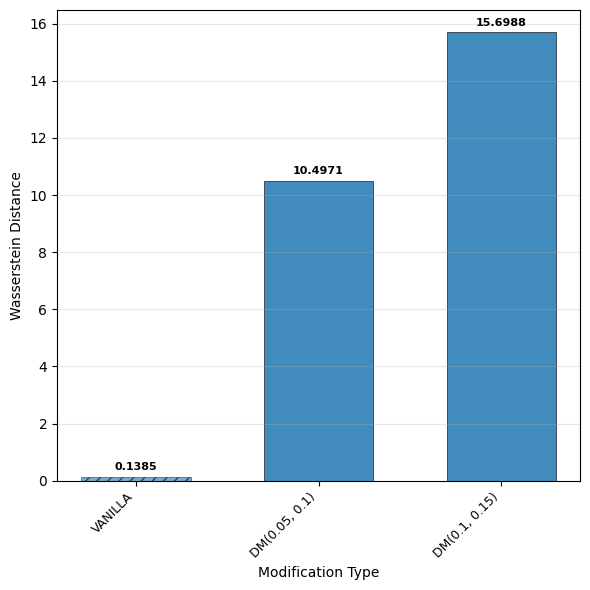}
        \caption{Wasserstein distances between training and synthetic datasets.}
        \label{fig:DM draw back image1}
    \end{subfigure}
    \hfill
\begin{subfigure}[t]{0.22\textwidth}
        \includegraphics[width=\linewidth]{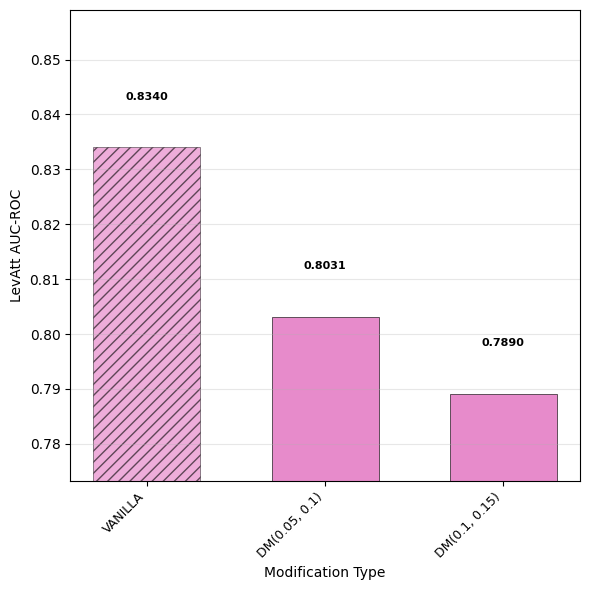}
        \caption{LevAtt AUC-ROC of synthetic datasets evaluated with an equal number of training and non-training records.}
        \label{fig:DM draw back image2}
    \end{subfigure}
    \caption{Privacy–fidelity comparison of DM on RealTabFormer synthetic data from the simulated dataset (Section \ref{sub:limits}). Vanilla corresponds to plain sampling without protection. Panel (a) reports Wasserstein distances; panel (b) shows LevAtt AUC-ROC. While DM is able to induce reductions in privacy leakage, the resulting synthetic data are of low fidelity.}
    \label{fig:DM draw back}
    \Description{Two-panel figure comparing privacy and fidelity metrics for differential privacy mechanisms applied to RealTabFormer. Left panel shows Wasserstein distances increasing with stronger privacy protection, indicating reduced data fidelity. Right panel shows LevAtt AUC-ROC decreasing with privacy protection, demonstrating the privacy-fidelity tradeoff where DM reduces privacy leakage at the cost of data quality.}
\end{figure}

Overall, we find that while effective in reducing privacy leakage, DM suffers large fidelity costs. In Figure~\ref{fig:DM draw back}, we applied DM to the simulation dataset, which is highly vulnerable to LevAtt (showing an MIA AUC-ROC above 83\% without any protection). As we increased the amount of noise injected into the synthetic data, LevAtt's AUC-ROC decreased. However, the fidelity gap measured by Wasserstein between the original training data and the noised synthetic data rose sharply. This sharp trade-off stems from the simulation dataset's small dynamic range and low variance - even small amounts of noise push values into out-of-domain regions. This illustrates that while DM is convenient- it can be applied post-hoc to any synthetic dataset- it is not an elegant protection mechanism: its agnostic approach to noise injection cannot respect the structural constraints that make synthetic data useful. 

\begin{figure*}[t]
    \centering
    \begin{subfigure}{0.45\textwidth}
        \includegraphics[width=\linewidth]{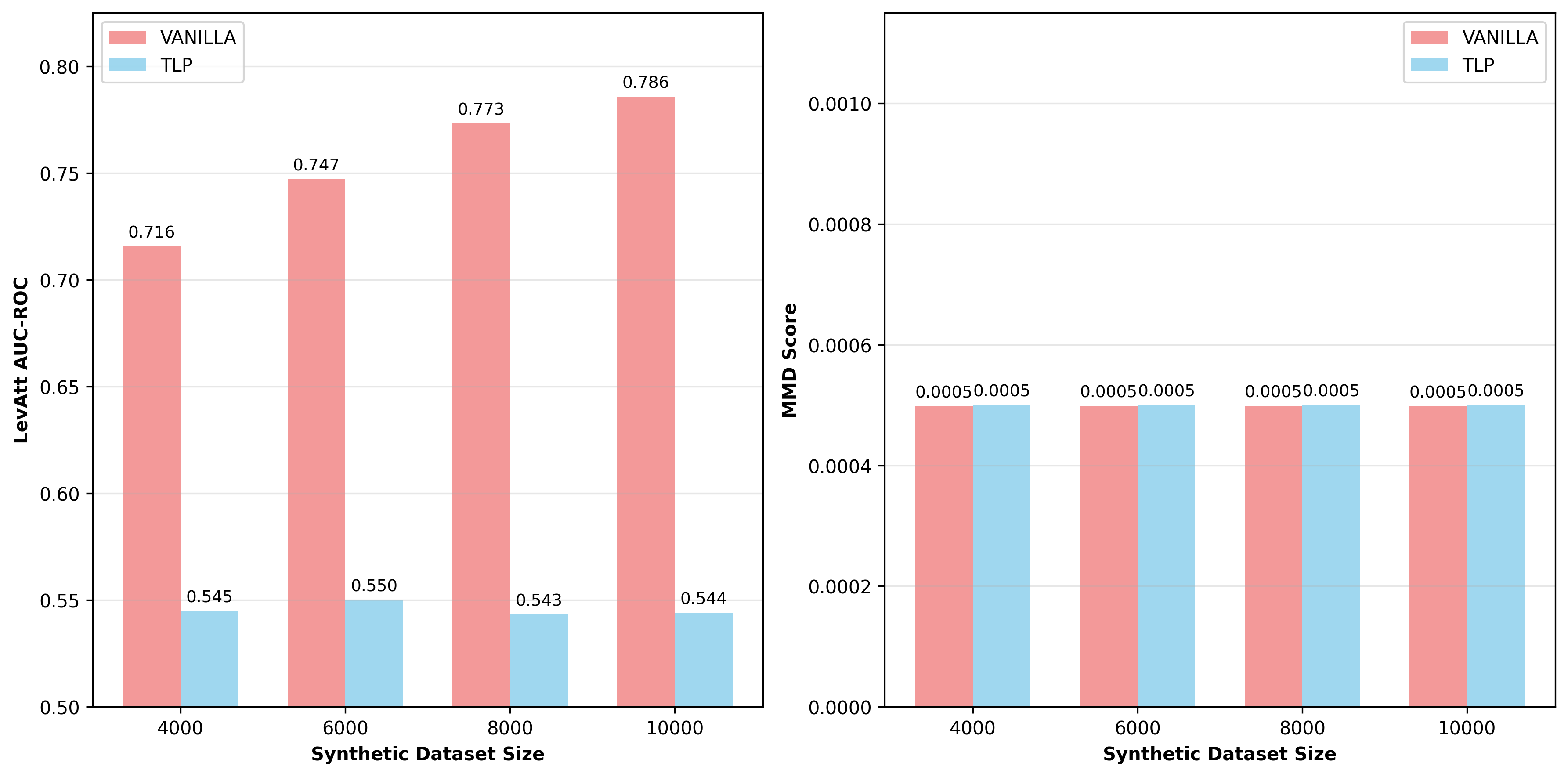}
        \caption{Simulation Dataset- LevAtt AUC and MMD}
        \label{fig:combined_simulation}
    \end{subfigure}
    \hfill
    \begin{subfigure}{0.45\textwidth}
        \includegraphics[width=\linewidth]{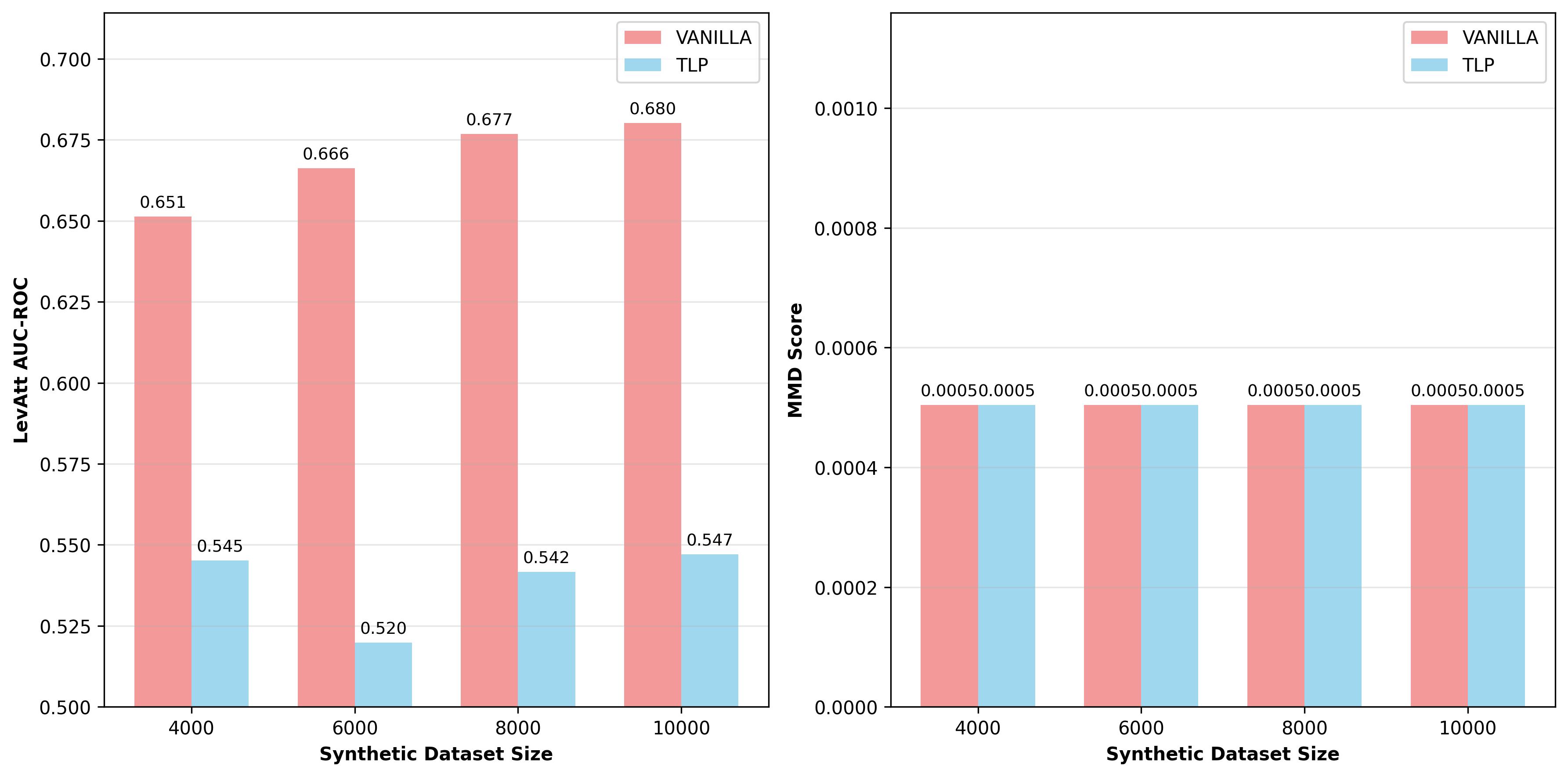}
        \caption{CASP Dataset- LevAtt AUC and MMD}
        \label{fig:combined_casp}
    \end{subfigure}
    
    \vspace{0.3cm}
    
    \begin{subfigure}{0.45\textwidth}
        \includegraphics[width=\linewidth]{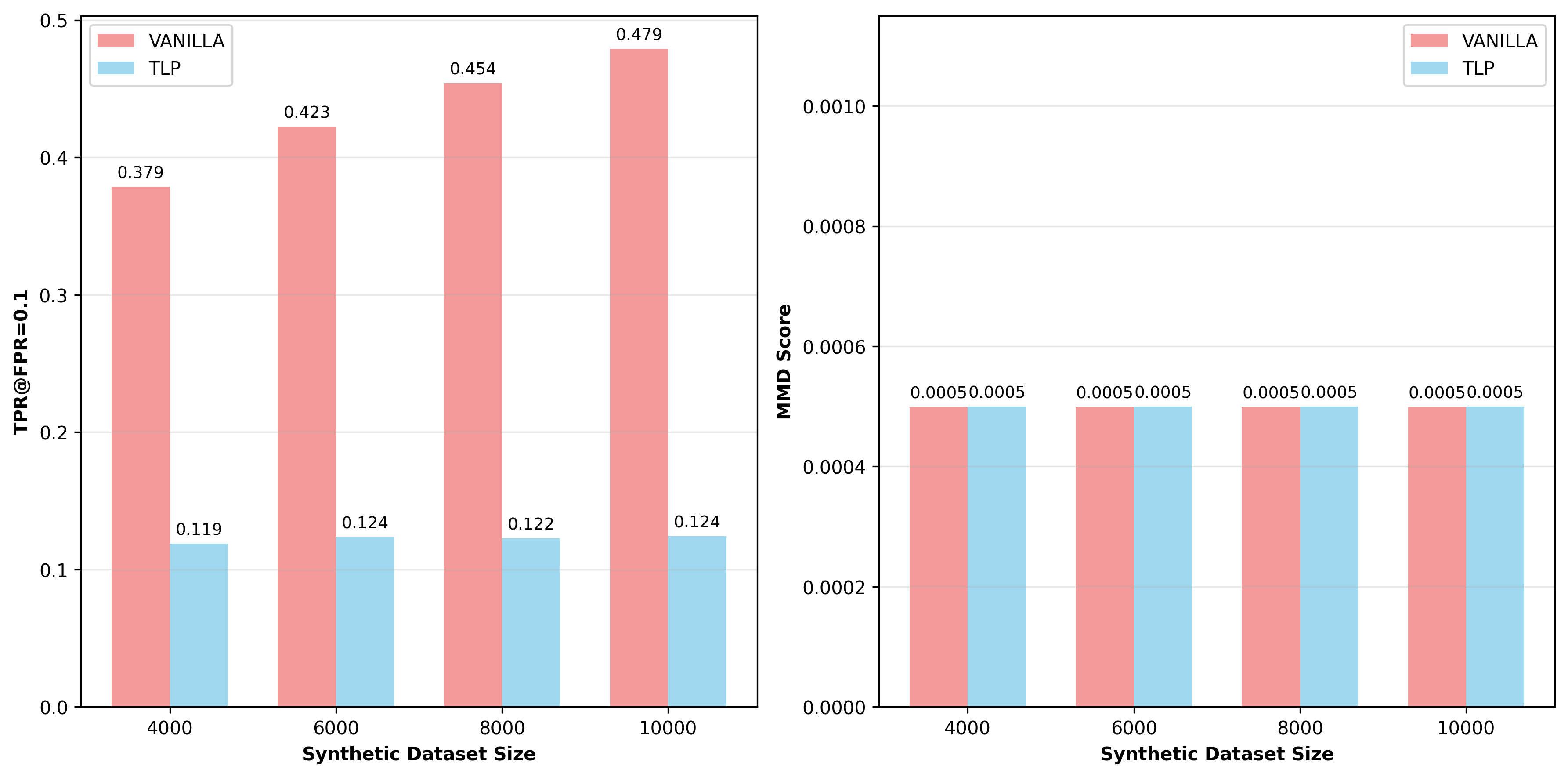}
        \caption{Simulation Dataset- LevAtt TPR@FPR=0.1 and MMD}
        \label{fig:combined_simulation_tpr}
    \end{subfigure}
    \hfill
    \begin{subfigure}{0.45\textwidth}
        \includegraphics[width=\linewidth]{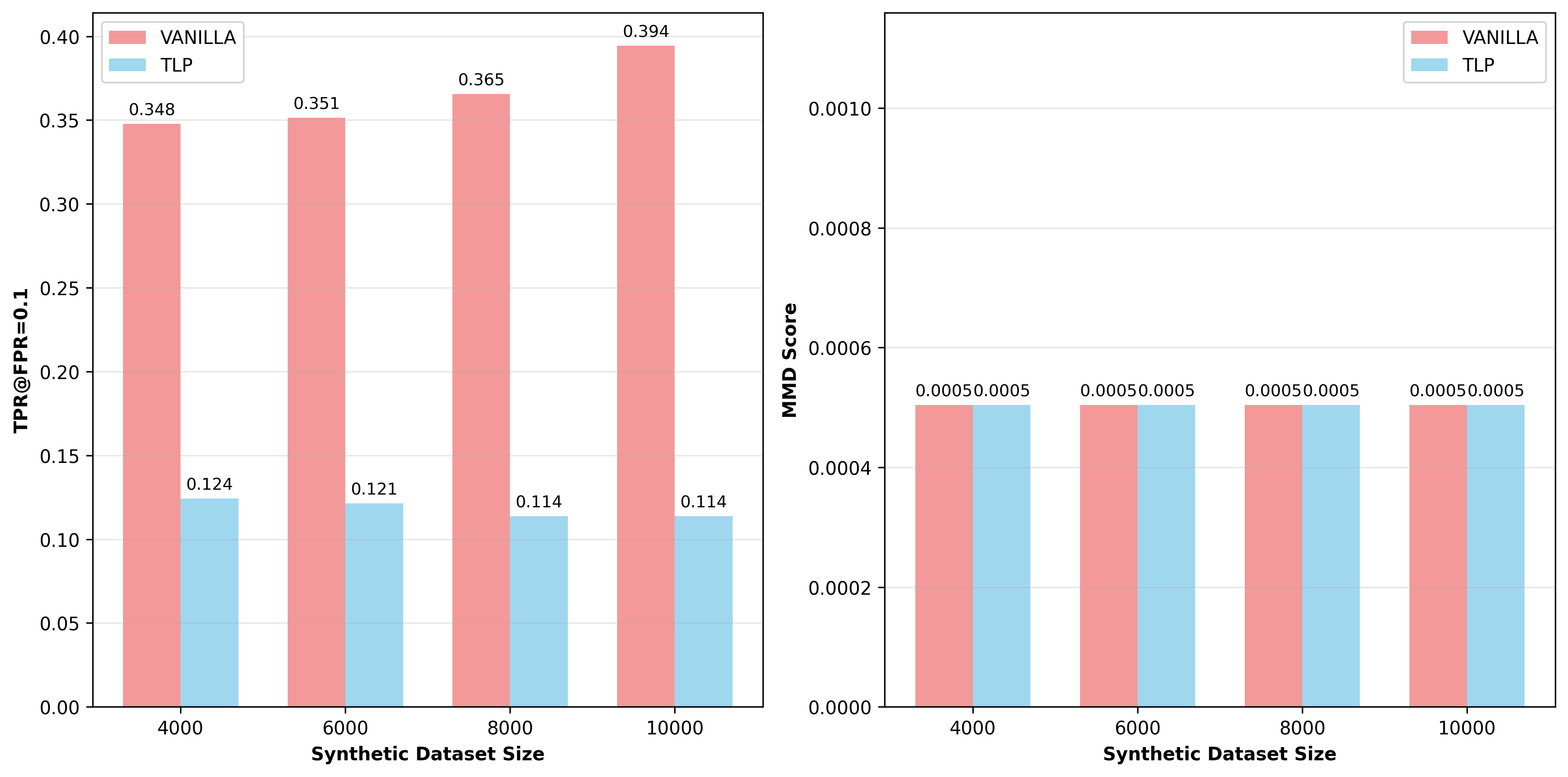}
        \caption{CASP Dataset- LevAtt TPR@FPR=0.1 and MMD}
        \label{fig:combined_casp_tpr}
    \end{subfigure}
    
    \caption{Privacy–fidelity trade-off of TLP on RealTabFormer synthetic data. We plot the AUC and Maximum Mean Discrepancy (MMD) for Vanilla and TLP-protected RealTabFormer models in (a)-(b). For both simulation (Section \ref{sub:limits}) and CASP datasets, TLP consistently reduces privacy leakage (AUC-ROC to $\sim$55\%) while creating synthetic data that matches the MMD of unmodified Vanilla data at various sizes. This trend holds as well for TPR@FPR=0.1 in (c)-(d) where TLP successfully reduces the TPR@FPR=0.1 to below 0.125 without loss to fidelity.}
    \Description{Four-panel comparison showing privacy-fidelity tradeoffs for TLP versus Vanilla RealTabFormer. Panels (a) and (b) display dual-axis plots with LevAtt AUC-ROC (left axis) and MMD (right axis) across different synthetic dataset sizes for Simulation and CASP datasets. TLP reduces AUC to approximately 55\% while maintaining similar MMD values to Vanilla. Panels (c) and (d) show similar dual-axis plots using TPR at FPR equals 0.1 metric, where TLP reduces TPR below 12.5 percent without degrading data fidelity as measured by MMD.}
    \label{fig:combined_privacy_utility}
\end{figure*}

 On the other hand, TLP- when equipped with a tuned tendency parameter $t$-controllably reduces attack efficacy while preserving the fidelity between the processed synthetic data and the training data. In our experiment, we begin with a small $t$ and evaluate its privacy protection effect. If the resulting synthetic data does not meet the privacy threshold (LevAtt AUC--ROC below $0.55$ or TPR below $0.125$ at FPR $=0.1$), we increment $t$ and repeat the procedure, stopping once we identify the smallest $t$ that satisfies the criterion. Across both the highly unprivate simulation setting and the CASP dataset, this tuned TLP reduces LevAtt's AUC from as high as $0.79$ to $0.55$ and drives the TPR@FPR=0.1 from $0.48$ to below $0.125$, all while incurring virtually no penalty in the Maximum Mean Discrepancy of the TLP-generated data across all synthetic data sizes (see Figure \ref{fig:combined_privacy_utility}).

TLP not only preserves the fidelity of the training dataset while meeting the privacy threshold, but also maintains the downstream utility of the CASP dataset. To evaluate this, we train XGBoost models on three versions of the data: real training subsets, vanilla synthetic data generated without protection, and TLP-processed synthetic data that satisfies the privacy requirement (LevAtt AUC-ROC below $0.55$ or TPR below $0.125$ at FPR $=0.1$). For each training size, model performance is assessed on the same held-out real test set using RMSE, and we observe that the utility degradation remains within a reasonable range (see Figure \ref{fig:TLP good Utility on CASP}).

% The DM faces a great challenge when the bandwidth of the dataset is relatively small. The most straightforward example is a dataset containing a column that has low variance. In such a case, modifying a digit could greatly pull the number in a region of low density. Another example of a small-bandwidth dataset is when all values in a given column lie within a restricted range. In such cases, modifying a digit can easily move the value outside its natural support. In general, it is challenging to design one comprehensive algorithm that assigns probability to each value from $[p_{min}, p_{max}]$ for the DM that could handle all restrictions in various datasets. To test the limitation of DM, we construct a simulated dataset with 20 columns, each independently sampled from a Gaussian distribution with mean 300 and standard deviation 5. A RealTabFormer model is trained on this dataset. At inference time, we generate synthetic datasets of the same size as the training data using either DM or the vanilla inference procedure without any protection mechanism. As shown in Figure \ref{fig:DM draw back}, DM substantially increases the Wasserstein distance, while only slightly reducing the AUC-ROC of our Levenshtein-distance-based MIA attack.

\section{Discussion}
To contextualize our findings and their implications, we discuss LevAtt’s broader significance, evaluate the effectiveness of proposed defenses, examine what LLMs truly learn from tabular data, and outline limitations and future research directions.
\subsection{LevAtt and the Privacy of LLM-Based Synthetic Data Generation}
LevAtt reveals that LLM-based tabular data generation is uniquely unsafe relative to conventional deep learning approaches. While being a simple string-similarity attack operating under an extremely restrictive threat model, LevAtt uncovers that state-of-the-art ICL models can catastrophically leak training membership by copying sequential patterns of digits and text from prompt exemplar examples. This phenomenon was further demonstrated in SFT-generators, revealing that the base implementation of RealTabFormer—a method recognized for its privacy-preserving capabilities—was susceptible to significant privacy leakage. In contrast, conventional tabular generators such as CT-GAN and TVAE were resistant to string-based attacks, and LevAtt exhibited low correlation with feature-space oriented MIAs.

This vulnerability stems from fundamental differences in how LLMs generate synthetic data. Unlike GANs and VAEs that model joint distributions directly in the feature space, LLMs decompose generation into sequential token prediction, where each digit is conditioned on previously generated values. This autoregressive process creates opportunities for the model to reproduce memorized digit sequences, particularly when training data contains repeated patterns, long numeric strings, or low-variance columns—structural characteristics common in real-world tabular datasets. These results highlight that autoregressive token-based generation in LLMs exposes a distinct attack surface not shared by other deep learning architectures, emphasizing the need for dedicated privacy audits in this domain.
\subsection{Sampling Defenses for LLMs}

While highly deployable and powerful, LevAtt can be defeated. In this work, we proposed an intuitive post-hoc defense called Digit Modifier (DM), which alters digits after generation. However, we found that DM failed to preserve the fidelity of the synthetic dataset. In contrast, our Tendency-based Logit Processor (TLP) successfully reduced LevAtt's AUC-ROC to below 55\% and TPR to below 12.5\% at FPR=10\% across both simulation and real-world datasets, while maintaining Maximum Mean Discrepancy nearly identical to vanilla generation. DM, despite achieving similar privacy improvements, incurred substantial fidelity costs, with Wasserstein distances increasing sharply as noise injection intensified.

The key advantage of TLP lies in its integration with the generation process itself. By perturbing logits during inference, TLP allows the model to maintain coherent dependencies across features while strategically introducing uncertainty in high-confidence digit predictions. DM, operating post-hoc, must blindly flip digits without access to the model's learned correlations, making it difficult to balance privacy protection with fidelity preservation. This is particularly problematic for datasets with narrow distributions or tight inter-feature dependencies, where even minor perturbations can push synthetic samples into low-density regions. TLP can be appended to any open-source LLM through the Hugging Face API, effectively controlling privacy leakage by smoothing the logits of digits with disproportionately high probabilities without substantially altering the statistical fidelity of the resulting synthetic data.

\subsection{Differential Privacy as Defense}
\begin{figure}[t]
    \centering
    \includegraphics[width=.75\linewidth]{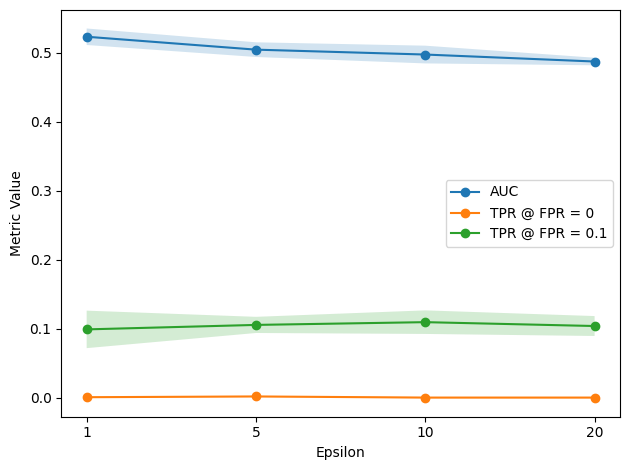}
    \caption{LevAtt performance on the Diabetes dataset at various $\epsilon$ levels for DP2Stage. }
    \label{fig:ep_ab}
    \Description{Line Graph of LevAtt performance on the Diabetes dataset at various $\epsilon$ levels for DP2Stage.}
\end{figure}

To mitigate the privacy risks exposed by LevAtt, we evaluate the effectiveness of implementing Differential Privacy (DP) during the training process. Our results indicate that across all datasets (See Table~\ref{tab:sft-levatt}, DP2Stage successfully defeats LevAtt. Indeed, we plot LevAtt's performance on the Diabetes dataset at varying levels of $\epsilon$ from DP2Stage in Figure~\ref{fig:ep_ab} and further find that all $\epsilon$ levels protect against LevAtt. However, this defense at training time comes at a substantial cost to data fidelity and utility as Table~\ref{tab:lev_fidelity_results} shows that DP2Stage had the highest Maximum Mean Discrepancy value and the worst ML AUC over all other models.

The application of DP in the context of Large Language Models (LLMs) faces two critical hurdles: first, DP methods are fundamentally incompatible with In-Context Learning (ICL), as ICL lacks a traditional training stage to privatize, necessitating our defenses at sample time instead. Second, growing evidence suggests LLMs are often pre-trained on common open-source benchmarking datasets \cite{bordt2024elephants}; if a model's weights have already been exposed to the target data prior to the private training stage, the technical validity of the DP guarantee is undermined. While DP2Stage offers superior protection against string-matching attacks like LevAtt, the trade-off between privacy and utility remains a primary bottleneck for high-fidelity tabular generation. 
\subsection{On the Nature of LLM-Based Tabular Learning}

While TLP provides an effective defense against LevAtt, the need for such inference-time mitigation raises a broader question: \textit{What are these models exactly learning?} A prevailing assumption is that LLM-based tabular generators approximate the joint distribution of the training set $T$ through sequential string modeling, similar to conventional generative approaches. However, our findings suggest that in some cases these models behave more like perturbation mechanisms, producing outputs that closely resemble training examples with only minor modifications. This behavior naturally yields high fidelity and downstream utility but does so precisely because of its proximity to the training data—thereby exposing the system to privacy leakage. 

This question is not merely theoretical—our experimental results provide concrete evidence of memorization-like behavior. The perfect membership classification achieved by LevAtt on some TabPFN-V2 runs, combined with the scaling of privacy leakage with model size, suggests that larger models increasingly rely on storing and retrieving training examples rather than abstracting general patterns. Furthermore, our finding that privacy leakage increases with both the volume of synthetic data generated (Figure \ref{fig:synth_size}) and the length of digit sequences in training data (Figure \ref{fig:digit_count}) aligns more closely with a retrieval mechanism than with principled distribution learning.

This question echoes similar debates in natural language processing, where LLMs demonstrably memorize training data under certain conditions \cite{carlini2021extractingtrainingdatalarge}. However, tabular data may be particularly susceptible to memorization due to its rigid structure and lack of paraphrase. In natural language, identical semantic content can be expressed in countless surface forms, creating ambiguity about whether a model has memorized a specific sentence or learned underlying concepts. Tabular data offers no such ambiguity—a sequence of digits either matches or does not. This rigidity may push LLM-based tabular generators toward memorization even when natural language applications achieve genuine generalization. Understanding when LLMs genuinely learn tabular distributions versus when they rely on approximate memorization remains an important direction for future work.

\subsection{Limitations}
While LevAtt and our proposed defenses demonstrate significant privacy vulnerabilities and mitigation strategies, several limitations warrant discussion.

First, LevAtt operates under a conservative No-box threat model, which—while demonstrating that privacy leakage occurs even under minimal adversarial assumptions—likely underestimates the true privacy risk in other possible scenarios. More permissive threat models that grant adversaries knowledge of model implementation details, training hyperparameters, or access to reference datasets would enable substantially more powerful attacks. For instance, knowledge of the specific tokenization scheme or access to auxiliary data could allow adversaries to adapt existing LLM-based membership inference techniques from natural language to tabular domains, potentially achieving even higher attack success rates than those reported here.

Second, while DM and TLP effectively reduce LevAtt's success rate, neither method provides formal privacy guarantees comparable to differential privacy. Both defenses are empirically validated against a specific attack rather than offering provable protection against arbitrary privacy audits. Consequently, more sophisticated attacks—particularly those that exploit model properties beyond string similarity—may successfully defeat these defenses \citep{mattern-etal-2023-membership, galli-etal-2024-noisy,10.5555/3737916.3742206}.

Third, we do not address adaptive adversaries who are aware that protective mechanisms are in place. An adversary with knowledge of TLP deployment, for example, might develop attacks that specifically target the statistical artifacts introduced by logit perturbation, or exploit correlations that remain intact despite digit modifications. Such adaptive strategies could potentially circumvent our defenses, highlighting the need for more robust, defense-aware attack evaluations and iterative improvement of protection mechanisms.

Finally, despite its effectiveness, LevAtt’s performance is contingent upon the synthetic generator’s adherence to a consistent string representation of the underlying data. Because Levenshtein distance operates at the character level, it can be sensitive to formatting discrepancies; for instance, if a model reproduces a memorized float like "17.5" as "17.500," the resulting edit distance may obscure the underlying membership signal despite the numerical values being identical. Furthermore, while the models in this study do not utilize open text, such data types present a unique challenge for string-based attacks. Unlike the rigid schemas of numerical or categorical fields, open text allows for semantic paraphrasing where the same information can be expressed through entirely different character sequences. In such instances, LevAtt might fail to detect memorized information if the surface form is altered, suggesting that extending this framework to unstructured text would require augmenting character-level metrics with semantic similarity measures.
\subsection{Future Work}
Our findings open several avenues for future research aimed at both strengthening the understanding of privacy leakage in LLM-based tabular data generators and developing more robust protective mechanisms.

A first direction is the exploration of richer threat models beyond the No-box setting. While our results show that even minimally informed adversaries can perform highly accurate membership inference, more permissive models may reveal additional vulnerabilities. Future work could investigate attacks that leverage model internals, surrogate data, or tokenization details, and evaluate how such enhanced adversarial capabilities affect memorization dynamics in structured data domains. This includes adapting or extending gradient-based or embedding-space attacks from NLP to tabular contexts, potentially uncovering deeper patterns of leakage.

Second, our proposed defenses—while effective against LevAtt—lack provable privacy guarantees. Future research should aim to establish theoretical foundations for privacy in LLM-based tabular generation, analogous to differential privacy frameworks used in classical synthetic data generation. One promising direction is designing logit- or representation-level perturbation mechanisms that preserve distributional fidelity while offering quantifiable bounds on memorization risk. Similarly, exploring how architectural choices, training objectives, or regularization schemes influence memorization could inform principled defense design.

\section{Conclusion}
In this work we introduce LevAtt, a No-box threat model MIA that exposes substantial privacy risk for in-context learning and supervised-finetuned LLM-based tabular data generators. LevAtt shows that LLMs are vulnerable to memorization from the structured, often duplicated patterns of tabular data. By attacking the string encodings of autoregressively generated tabular data, LevAtt finds unique adversarial signal compared to existing methods. While less restrictive threat models would likely lead to a better attack, we believe No-box carries a powerful message: an attack with the minimal assumptions reasonably possible for an adversary can perfectly classify training membership in state-of-the-art generators. Lastly propose two defenses against LevAtt, showing that Tendency-Based Logit Processor can effectively defeat LevAtt with minimal loss in synthetic data fidelity. Future research directions could involve developing even more powerful attacks under less restrictive threat models, finding more efficient and provable defenses, and studying mechanistically how LLMs learn and represent tabular distributions.

\section{Statement of Ethics}
The potential for adversaries to determine whether an individual’s data was included in the original dataset presents significant privacy risks, especially in fields such as healthcare, finance, and social sciences, where sensitive personal information is commonly used. Synthetic data that fails to sufficiently mask membership information could inadvertently enable re-identification. Although this work introduces a method for assessing such risks, its primary objective is to empower researchers and practitioners to perform more rigorous privacy evaluations before deploying synthetic datasets. We emphasize that adversarial approaches are essential for advancing the development of robust privacy-preserving systems.
\begin{acks}
    The authors used LLMs to assist in the programming of our experiments, designing data visualizations, making the codebase more user-friendly, finding clearer phrasings in our writing, and formatting Latex code. All revisions by LLMs were checked and validated by the authors.

    This work was financially supported by NSF – CNS (2247795), Office of Naval Research, (ONR N00014-22-1-2680), a gift from CISCO, and the National AI Research Resource (NAIRR-250187). 

\end{acks}
% \section{Acknowledgments}

% % \section{Reproducibility Statement}
% % Our code is available through the link provided on the abstract page. The main paper explains all algorithms in detail and provides the dataset and simulation descriptions, while model hyperparameters are reported in the appendix. 
\bibliographystyle{ACM-Reference-Format}
\bibliography{main.bib}

\appendix
\section{Appendix}
\label{app:section4}
\subsection{In-Context Learning details}
\label{sub:ICL-exp}

\subsubsection{Generator Details}
\begin{enumerate}
    \item \textbf{TabPFN Generation~\cite{hollmann2025accurate}}:
We followed the Prior Labs tutorial (\url{https://priorlabs.ai/tutorials/unsupervised/}) for unsupervised TabPFN. Each training split was loaded, shuffled, and divided into batches of 200 rows. Numeric features were cast to \texttt{float32}, while categorical variables were label-encoded (assigning unseen categories to --1). Columns with zero variance were removed prior to model fitting and reintroduced after sampling. For each batch, we fit TabPFN and generated synthetic rows using temperature $t=1.0$ across three random permutations. Outputs were decoded, constant columns reattached, batches concatenated, and finally truncated to match the size of the original dataset.
\item \textbf{LLaMA Generation~\cite{meta2024llama3_3_70b}}:
We used LLaMA 3.3 70B via the Groq API (\url{https://console.groq.com/docs/models}). Each training split was divided into batches of up to 32 rows, ensuring that all rows were fully included in the prompt. For each batch, we computed per-column summary statistics and serialized the data to CSV. We then queried \texttt{llama-3.3-70b\-versatile} with temperature $t=1.0$, requesting $N$ rows in JSON format. If outputs contained parse errors or incorrect row counts, we retried the generation up to five times. Valid generations were concatenated, truncated, or re-prompted as necessary, and validated for type and dimensional consistency.\footnote{LLaMA 3.3-70B failed on \textit{geographical-origin-of-music}, \textit{pumadyn32nh}, \textit{student-performance-por}, \textit{superconductivity}, and \textit{wave-energy} due to token limitations. TabPFN failed on \textit{geographical-origin-of-music} due to extreme dimensionality.}
\item \textbf{GPT-4o-mini~\cite{openai2024gpt4omini}}:
We applied the same prompting and inference pipeline as with LLaMA 3.3 70B. However, we used OpenAI's structured output API, defining the target format as a JSON schema with column names as keys and corresponding cell values as entries.
\end{enumerate}

\label{sec:prompt}

\begin{lstlisting}[basicstyle=\ttfamily\footnotesize, breaklines=true, frame=single, caption={ICL Prompt Template passed to Groq API}, label={lst:prompt}, float=t]
System role: You are a tabular synthetic data generation model.

Your goal is to produce data that mirrors the given examples in 
causal structure and feature/label distributions, 
while maximizing diversity.

Context: Leverage your in-context learning to generate realistic, 
diverse samples.

Output format: JSON.

Dataset name: {dataset_name}

Column names (in order): {col_names}

Summary statistics:
{summary_stats}

CSV of full data:
{data}

Please generate {batch_size} rows of synthetic data.

Treat the rightmost column as the target. Return only a JSON object:
{
  "synthetic_data": "<CSV string>"
}

Do not include any additional text.
\end{lstlisting}
\subsection{Black-Box MIAs}
We apply feature space based black box MIAs in the same experimental design as LevAtt. Here, we treat the exemplar observations as the membership class and holdout data as non-membership. Before calling the attacks, we scale continuous and one-hot encode categorical variables for DCR and MC. As Kernel Density Estimation often fails to converge under one-hot encoded categorical variables we instead ordinal encode them. All implementations of these attacks are through the Synth-MIA library \cite{Synth-MIA}.
\begin{enumerate}

    \item \textbf{Distance to Closest Record (DCR)} \cite{ganleaks}: DCR is a black-box attack that scores test data based on a score of the Euclidean distance to the nearest neighbor in the synthetic dataset. 
    
    \item \textbf{MC} \cite{Hilprecht2019MonteCA}: MC is based on counting the number of observations in the synthetic dataset that fall into the neighborhood of a test point (Monte Carlo Integration). However, this method does not consider a reference dataset, and the choice of distance metric for defining a neighborhood is a non-trivial hyperparameter to tune.
        
    \item \textbf{Density Estimate} \cite{Hilprecht2019MonteCA, vanbreugel2023membership}: Density Estimate follows a similar strategy as MC, but rather than using a Monte Carlo approximation of density, instead uses a Kernel Density Estimator (KDE). The idea is that a test observation that is a training observation will have a higher density estimate on a KDE fit over a synthetic dataset.
    \end{enumerate}
\subsection{SFT-Datasets}
\begin{table}[ht]
\begin{tabular}{lcc}
\hline
\textbf{Dataset} & \textbf{\# Instances} & \textbf{\# Features} \\
\hline
\href{https://www.openml.org/search?type=data&sort=runs&id=183&status=active}{Abalone} (\href{https://www.openml.org}{OpenML}) & 4,177 & 9 \\
\href{https://www.openml.org/search?type=data&status=active&id=45578&sort=runs}{CA Housing} (\href{https://www.openml.org}{OpenML}) & 20,640 & 9 \\
\href{https://www.openml.org/search?type=data&status=active&id=42903}{CASP} (\href{https://www.openml.org}{OpenML}) & 45,730 & 9 \\
\href{https://archive.ics.uci.edu/dataset/34/diabetes3}{Diabetes} (\href{https://scikit-learn.org/stable/modules/generated/sklearn.datasets.load_diabetes.html}{Sklearn}) & 412 & 10 \\
\href{https://archive.ics.uci.edu/dataset/198/steel+plates+faults}{Steel Plates Faults} (\href{https://archive.ics.uci.edu}{UCI}) & 1,941 & 27 \\
\hline
\end{tabular}
\label{tab:dataset_summary}
\end{table}

\subsection{SFT Model Details}
We modify the original RealTabFormer implementation to use LLaMA 3.2 (1B, 3B) \citep{grattafiori2024llama3herdmodels}, Qwen2.5-3B \citep{qwen2025qwen25technicalreport}, and Mistral v0.3 7B \cite{jiang2023mistral7b}. Here, we follow RealTabFormer's base training and sampling hyperparameters in Table \ref{tab:rtfhyp}. To SFT GREAT, we also use its original implementation of which the base hyperparameters can be found in Table \ref{tab:greathyp}. For CT-GAN and TVAE, we use the default hyperparameters and implementation found in Synthcity \citep{synthcity}.

\begin{table}[h!]
\centering
\begin{tabular}{|l|l|}
\hline
\textbf{Hyperparameter} & \textbf{Default Value} \\ \hline
\texttt{epochs} & 1000 \\ \hline
\texttt{batch\_size} & 8 \\ \hline
\texttt{train\_size} & 1 \\ \hline
\texttt{output\_max\_length} & 512 \\ \hline
\texttt{early\_stopping\_patience} & 5 \\ \hline
\texttt{early\_stopping\_threshold} & 0 \\ \hline
\texttt{mask\_rate} & 0 \\ \hline
\texttt{numeric\_nparts} & 1 \\ \hline
\texttt{numeric\_precision} & 4 \\ \hline
\texttt{numeric\_max\_len} & 10 \\ \hline
\end{tabular}
\caption{Numeric hyperparameters of REaLTabFormer.}
\label{tab:rtfhyp}
\end{table}

\begin{table}[h!]
\centering
\begin{tabular}{|l|l|}
\hline
\textbf{Hyperparameter} & \textbf{Default Value} \\ \hline
\texttt{epochs} & 100 \\ \hline
\texttt{batch\_size} & 8 \\ \hline
\texttt{float\_precision} & None \\ \hline
\texttt{temperature} & 0.7 \\ \hline
\texttt{top-k sampling} & 100 \\ \hline
\texttt{max\_length} & 100 \\ \hline
\end{tabular}
\caption{GReaT training and sampling hyperparameters.}
\label{tab:greathyp}

\end{table}

\subsection{Simulated Data Generation Details}
For Figure \ref{fig:digit_count}, we initialize a Multivariate Gaussian of N(1e10,1e9). This ensures that there are up to 10 digits for a column as RealTabFormer can struggle to process exceptionally long decimal strings. We then sample 10,000 training and holdout rows for our experiment. At each level, we add additional columns to the training, holdout, and therefore synthetic dataset observation sequence lengths increase by 10 digits for each column.

\section{More Details on Defenses}
\label{sub:More Details On Defenses}
\textbf{Digit Substitution Rule in DM:}
For the Digit Modifier (DM), each selected digit is perturbed using a simple increment rule. Specifically, a digit \(d \in \{0,1,\ldots,9\}\) is replaced by \((d+1) \bmod 10\). Thus, \(0\) becomes \(1\), \(1\) becomes \(2\), and so on, with \(9\) wrapping around to \(0\). This cyclic increment operation provides a minimal yet consistent perturbation to each affected digit, ensuring that the modified values remain close to their originals while still introducing controlled randomness. 
% \vspace{0.2cm}

\noindent\textbf{Handling Invalid Logits in ReaLTabFormer:}
In ReaLTabFormer, some logits take the value \(-\infty\) or become \texttt{NaN} due to masking constraints in the tabular structure. These values arise when certain tokens are structurally disallowed, leading to zero-probability entries after masking. Since such logits cannot be meaningfully transformed by TLP, we simply ignore them: TLP is applied only to finite-valued logits, while any \(-\infty\) or \texttt{NaN} logits are left unchanged to preserve the validity of the generator’s masking logic.

% \vspace{0.2cm}
\noindent\textbf{Why TLP Is Applied Only at Inference Time:}
The Tendency-based Logit Processor (TLP) is designed exclusively for inference-time perturbations. Applying TLP during training severely disrupts the optimization dynamics: the modified logits distort the gradient signal, causing the loss to decrease extremely slowly and preventing the model from converging in a reasonable number of steps. In effect, the model must simultaneously learn both the underlying data distribution and the perturbation induced by TLP, which dramatically complicates training. By restricting TLP to inference time, we preserve stable training behavior while still introducing controlled variability into the generated samples.

\noindent\textbf{Role of the Stabilization Constant \(\varepsilon\).}
The shifted min–max scaling used in TLP requires a small positive constant \(\varepsilon\) to ensure numerical stability. In cases where all logits in a column share the same value, we have \(M_l = m_l\), causing the denominator \(M_l - m_l\) in the scaling expression to become zero. Without a stabilizing term, this would result in undefined values or \texttt{NaN} outputs after normalization. By adding a small \(\varepsilon > 0\), we guarantee that the denominator remains strictly positive, preventing division-by-zero errors and ensuring that the scaled logits remain well defined. This safeguard is essential for applying TLP robustly, especially when the generator outputs nearly constant logits due to structural constraints in the data.

\section{Additional Tables and Figures} \label{sec:tables}

\begin{table*}
\caption{ICL Experiment Dataset Characteristics}
\label{tab:ctr23_metadata}
\begin{tabular}{lrrr}
\toprule
Dataset Name & Num. Numeric & Num. Categorical & Instances \\
\midrule
\text{Moneyball} & 9 & 6 & 1232 \\
\text{QSAR\_fish\_toxicity} & 7 & 0 & 908 \\
\text{abalone} & 8 & 1 & 4177 \\
\text{airfoil\_self\_noise} & 6 & 0 & 1503 \\
\text{auction\_verification} & 6 & 2 & 2043 \\
\text{brazilian\_houses} & 6 & 4 & 10692 \\
\text{california\_housing} & 9 & 0 & 20640 \\
\text{cars} & 18 & 0 & 804 \\
\text{concrete\_compressive\_strength} & 9 & 0 & 1030 \\
\text{cps88wages} & 3 & 4 & 28155 \\
\text{cpu\_activity} & 22 & 0 & 8192 \\
\text{diamonds} & 7 & 3 & 53940 \\
\text{energy\_efficiency} & 9 & 0 & 768 \\
\text{fifa} & 28 & 1 & 19178 \\
\text{forest\_fires} & 11 & 0 & 517 \\
\text{fps\_benchmark} & 31 & 13 & 24624 \\
\text{geographical\_origin\_of\_music} & 117 & 0 & 1059 \\
\text{grid\_stability} & 13 & 0 & 10000 \\
\text{health\_insurance} & 5 & 7 & 22272 \\
\text{kin8nm} & 9 & 0 & 8192 \\
\text{kings\_county} & 18 & 4 & 21613 \\
\text{miami\_housing} & 16 & 0 & 13932 \\
\text{naval\_propulsion\_plant} & 15 & 0 & 11934 \\
\text{physiochemical\_protein} & 10 & 0 & 45730 \\
\text{pumadyn32nh} & 33 & 0 & 8192 \\
\text{red\_wine} & 12 & 0 & 1599 \\
\text{sarcos} & 22 & 0 & 48933 \\
\text{socmob} & 2 & 4 & 1156 \\
\text{solar\_flare} & 3 & 8 & 1066 \\
\text{space\_ga} & 7 & 0 & 3107 \\
\text{student\_performance\_por} & 14 & 17 & 649 \\
\text{superconductivity} & 82 & 0 & 21263 \\
\text{video\_transcoding} & 17 & 2 & 68784 \\
\text{wave\_energy} & 49 & 0 & 72000 \\
\text{white\_wine} & 12 & 0 & 4898 \\
\bottomrule
\end{tabular}
\end{table*}

\begin{table*}
\centering
\caption{SFT Experiment Dataset Characteristics}
\label{tab:sft_datasets}
\begin{tabular}{l c c c}
\hline
{Dataset Name} & {Num. Numeric} & {Num. Numeric}  & {Instances} \\ \hline
Housing          & 1                    & 9                                      & 20,640           \\
Diabetes         & 2                    & 7                                     & 768              \\
CASP             & 0                    & 10                                  & 45,730           \\
Faults           & 10                   & 24                                  & 1,941            \\
Abalone          & 2                    & 7                                  & 4,177            \\ \hline
\end{tabular}
\end{table*}
% \section{Statement of LLM Use}
% We used LLMs to assist in the programming of our experiments, designing data visualizations, making the codebase more user-friendly, finding clearer phrasings in our writing, and formatting Latex code. All additions by LLMs were checked by the authors.
\begin{table*}[]
\centering
\caption{Privacy and fidelity metrics across ICL generators and context size (Mean $\pm$ STD). Runs with fewer context example sizes demonstrate both higher risk to LevAtt and lower utilities.}
\label{tab:icl_n_size}
\begin{tabular}{llcccccc}
\toprule
Generator & Size & AUC & TPR@0 & TPR@0.1 & MMD & Wasserstein & ML AUC \\
\midrule

\multirow{3}{*}{LLaMA-3.3-70B}
& 32  & 0.67 $\pm$ 0.146 & 0.11 $\pm$ 0.246 & 0.34 $\pm$ 0.265 & 0.06 $\pm$ 0.016 & 0.65 $\pm$ 0.581 & 0.61 $\pm$ 0.113 \\
& 64  & 0.62 $\pm$ 0.134 & 0.07 $\pm$ 0.184 & 0.26 $\pm$ 0.237 & 0.03 $\pm$ 0.025 & 0.67 $\pm$ 0.831 & 0.62 $\pm$ 0.115 \\
& 128 & 0.60 $\pm$ 0.112 & 0.07 $\pm$ 0.172 & 0.24 $\pm$ 0.208 & 0.02 $\pm$ 0.005 & 4090.75 $\pm$ 38150.682 & 0.66 $\pm$ 0.123 \\
\midrule

\multirow{3}{*}{TabPFN-V2 }
& 32  & 0.59 $\pm$ 0.112 & 0.04 $\pm$ 0.176 & 0.20 $\pm$ 0.200 & 0.06 $\pm$ 0.006 & 1.12 $\pm$ 1.155 & 0.62 $\pm$ 0.103 \\
& 64  & 0.57 $\pm$ 0.111 & 0.04 $\pm$ 0.173 & 0.18 $\pm$ 0.207 & 0.03 $\pm$ 0.004 & 0.95 $\pm$ 1.100 & 0.68 $\pm$ 0.117 \\
& 128 & 0.56 $\pm$ 0.113 & 0.04 $\pm$ 0.172 & 0.17 $\pm$ 0.205 & 0.02 $\pm$ 0.003 & 0.81 $\pm$ 1.052 & 0.70 $\pm$ 0.115 \\
\midrule

\multirow{3}{*}{GPT-4o-mini}
& 32  & 0.55 $\pm$ 0.055 & 0.00 $\pm$ 0.009 & 0.15 $\pm$ 0.063 & 0.07 $\pm$ 0.019 & 14.45 $\pm$ 67.815 & 0.57 $\pm$ 0.087 \\
& 64  & 0.53 $\pm$ 0.045 & 0.00 $\pm$ 0.010 & 0.12 $\pm$ 0.040 & 0.04 $\pm$ 0.010 & 13486.80 $\pm$ 126698.02 & 0.59 $\pm$ 0.102 \\
& 128 & 0.52 $\pm$ 0.026 & 0.00 $\pm$ 0.003 & 0.11 $\pm$ 0.029 & 0.03 $\pm$ 0.106 & 72.58 $\pm$ 407.985 & 0.62 $\pm$ 0.099 \\
\bottomrule
\end{tabular}
\end{table*}

\begin{figure*}
    % \label{fig:power_roots}
    \centering
    \includegraphics[width=0.45\textwidth]{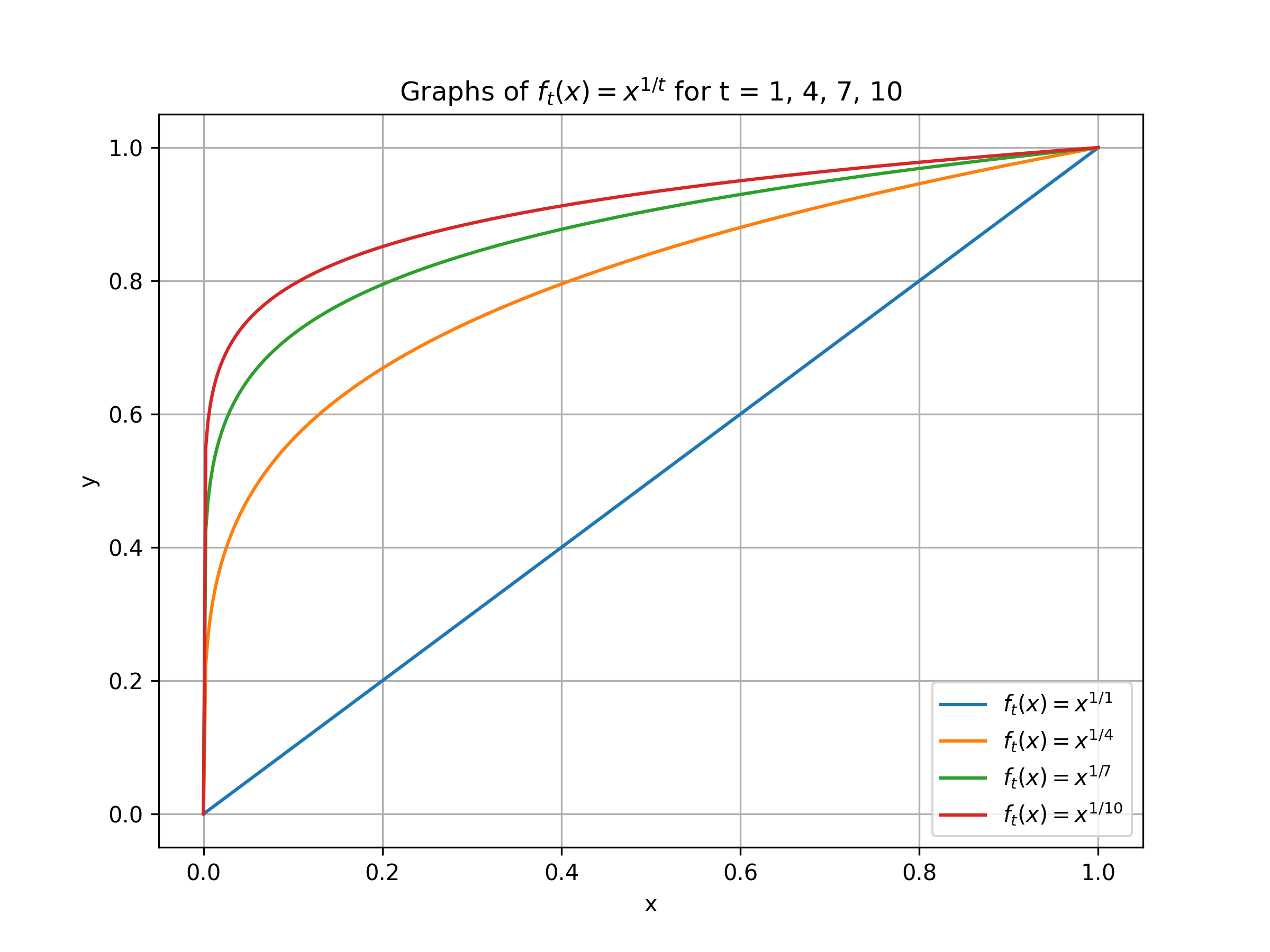}
    \caption{Visualization of the transformation function \( f_t(x) = x^{1/t} \) under varying values of \( t \). As \( t \) increases, the function becomes more concave, amplifying smaller logits proportionally more than larger ones. This behavior helps compress logit differences while preserving their ordering.}
    \Description{Graph showing multiple curves of the power root transformation function for different values of t. The curves start at the origin and increase concavely, with higher t values creating more pronounced concave shapes that amplify smaller input values more than larger ones.}
    \label{fig:power_roots}
\end{figure*}

\begin{figure*}
    \centering
    \begin{subfigure}{0.48\textwidth}
        \includegraphics[width=\linewidth, alt={Bar graph showing RMSE of CASP dataset XGBoost models}]{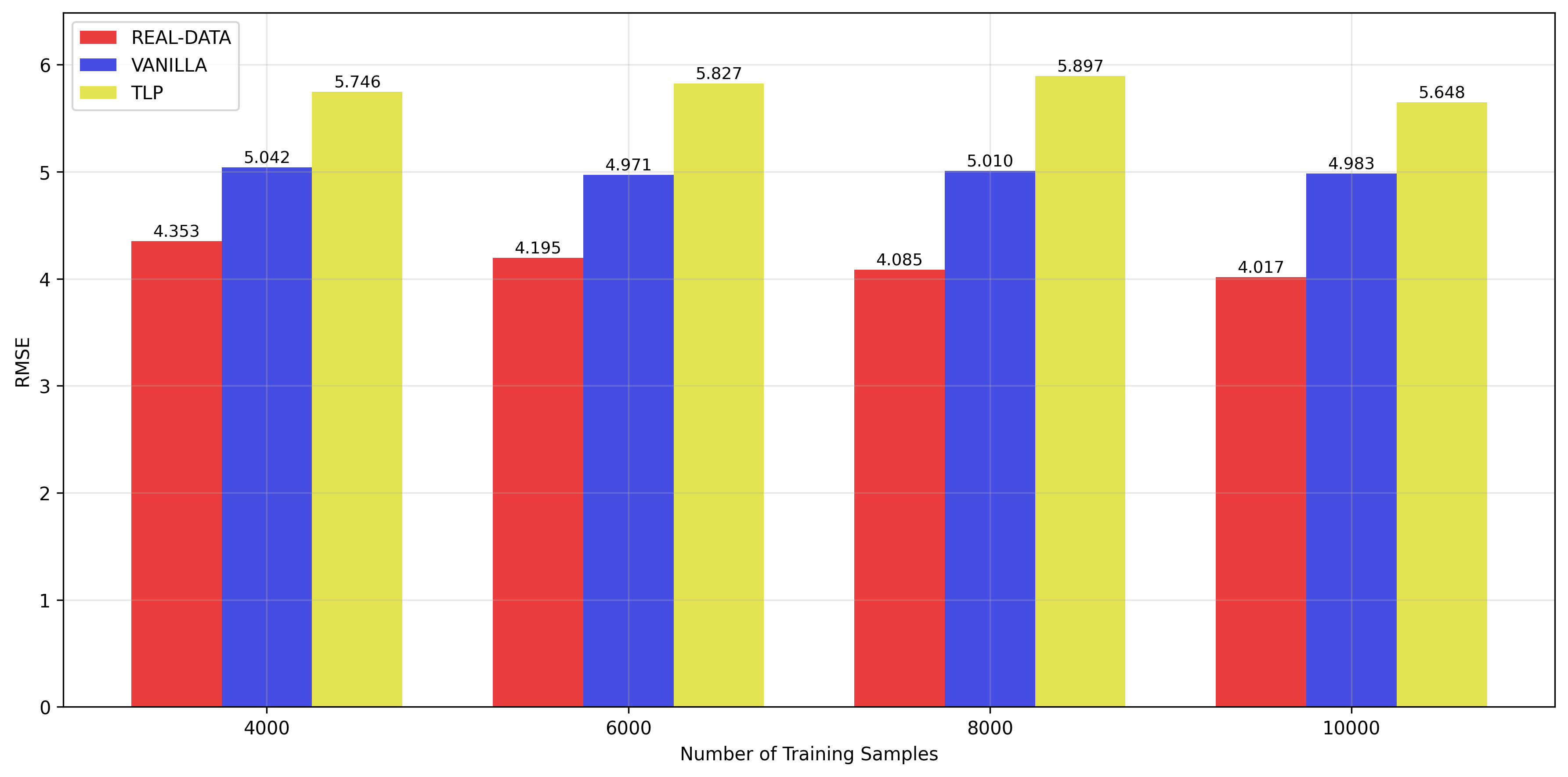}
        \caption{RMSE of CASP dataset XGBoost models where TLP-protected datasets have a LevAtt AUC of below 0.55.}
        \label{fig:CASPXG}
    \end{subfigure}
    \hfill
    \begin{subfigure}{0.48\textwidth}
        \includegraphics[width=\linewidth,alt={Bar graph showing RMSE of CASP dataset XGBoost models}]{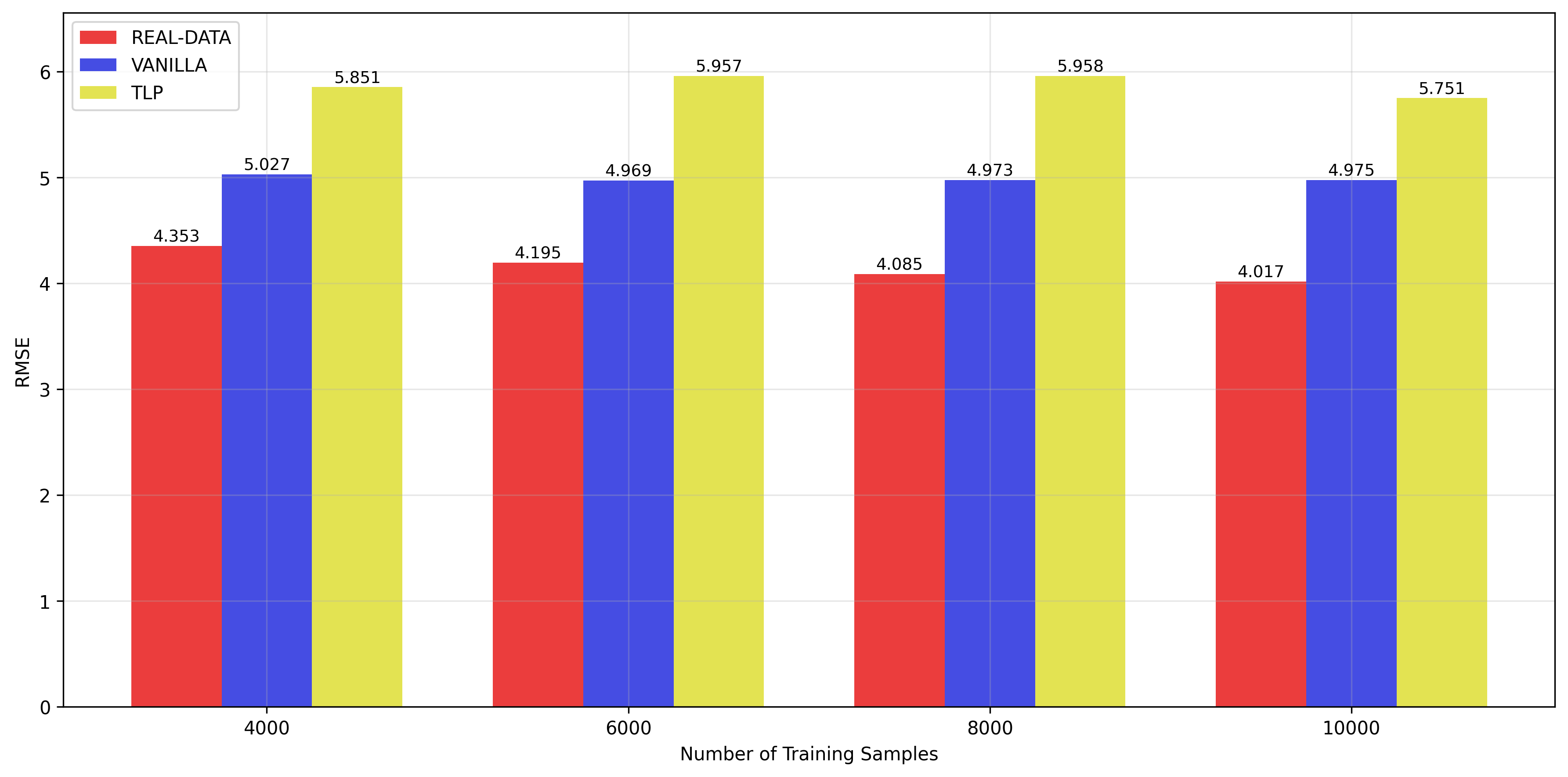}
        \caption{RMSE of CASP dataset XGBoost models where TLP-protected datasets have a LevAtt TPR@FPR=0.1 of below 0.125.}
        \label{fig:TLP good fidelity on simulation}
    \end{subfigure}
    \caption{Utility comparison of XGBoost models trained on real, vanilla synthetic, and TLP-protected synthetic data at various synthetic dataset (and thus privacy leakage) sizes. Real data achieves the lowest RMSE, vanilla synthetic shows moderate degradation, and TLP-protected data shows larger degradation. However, the gap between vanilla and TLP remains stable as training size increases, indicating that TLP provides a controllable privacy–utility trade-off even when stronger tendency levels are needed.}
    \Description{Figure showing utility comparison of XGBoost models trained on real, vanilla synthetic, and TLP-protected synthetic data across various dataset sizes.}
    \label{fig:TLP good Utility on CASP}
\end{figure*}

\end{document}